\documentclass[11pt]{article}

\usepackage[margin=1in]{geometry}
\usepackage[T1]{fontenc}
\usepackage[utf8]{inputenc}
\usepackage{lmodern}
\usepackage{microtype}
\usepackage{amsmath,amssymb,mathtools}
\usepackage{booktabs,longtable,tabularx,array,multirow}
\usepackage{graphicx}
\graphicspath{{figures/}}
\usepackage{xcolor}
\usepackage{enumitem}
\usepackage{authblk}
\usepackage{url}
\usepackage[hidelinks]{hyperref}
\usepackage{caption}
\usepackage{subcaption}
\usepackage{placeins}

\newcommand{\sys}{\textsc{QuantPits}}
\newcommand{\oomrl}{\textsc{OOM-RL}}

\title{\textbf{OOM-RL II: Reality Is an Oracle, Not a Debugger}\\
\large Provenance-Constrained Diagnosis in Continually Evolving Agent-Engineered Systems}

\author[1]{Kun Liu\thanks{Corresponding author: \texttt{ai@quantpits.com}}}
\author[1]{Liqun Chen}
\affil[1]{QuantPits.com}

\hypersetup{
  pdftitle={OOM-RL II: Reality Is an Oracle, Not a Debugger},
  pdfauthor={Kun Liu and Liqun Chen}
}
\date{}

\begin{document}
\maketitle

\begin{abstract}
Reality may establish that an outcome occurred without identifying which evolving procedure produced it or why. This distinction matters in production ML systems whose code, configuration, and artifacts change while external feedback accumulates. We examine it in a human-directed, agent-engineered quantitative trading system, using \emph{oracle} to mean an external source of realized outcomes rather than a complete correctness specification.

Across one year, the account gained and outperformed a broad market index, while annual alpha was not statistically distinguishable from zero under the main retrospective specification. Retrospectively selected subperiods include adverse relative performance and conditional candidate-level weakness under declared approximate references. Engineering records document changes during the episode, and complete recommendation-to-runtime binding is unavailable. The archive does not establish a common frozen instance or a unique cause.

The case motivates an \emph{outcome--diagnosis gap}: outcome evidence, evaluated-object identity, and causal explanation support distinct claims. We distinguish frozen instances, pre-specified adaptive procedures, and ad-hoc development; organize archive-relative claim identifiability and an evidence hierarchy; and propose a prospective production-binding protocol. An illustrative compatible-history example shows how factual binding can resolve a recommendation's referent without supplying its counterfactual effect. The protocol is proposed rather than prospectively validated. External feedback constrains outcome claims, while provenance and additional identification structure determine the resolution of diagnosis.
\end{abstract}


\section{Introduction}

An external outcome records what happened, but does not identify the internal procedure or mechanism responsible for it. This distinction becomes consequential when a production system receives feedback while its code, artifacts, configuration, and operational rules are being revised. The system evaluated today may differ from the procedure that produced yesterday's output.

SWE-bench provides repository issue-resolution tasks, while SWE-agent illustrates tool-mediated repository navigation, code editing, and program execution \cite{jimenez2024swebench,yang2024sweagent}. LLM agents can participate in production engineering through such activities. Their participation raises questions about the identity of the resulting system and the authority to change or deploy it. The identity problem is not unique to LLM agents: it also arises in conventional MLOps. Here agent-assisted engineering makes tasks, tool-mediated proposals, and acceptance decisions relevant parts of the longitudinal provenance chain.

This paper asks: \emph{what does an external outcome actually tell us about an evolving system?}

A financial loss is a stochastic outcome, not a built-in software correctness label. Even when a portfolio performs poorly relative to an explicit reference, the observed result does not name the faulty model, the faulty line of code, the faulty training mode, or the relevant environmental mechanism. The threshold defining ``failure,'' the comparator, and the analysis window must themselves be specified. Reality can therefore be authoritative about what happened while remaining ambiguous about why it happened.

The distinction is concrete in \sys{}, an agent-engineered production ML system. Here \emph{agent-engineered} means developed and revised with LLM-agent assistance under human-issued tasks, with human operational and deployment authority retained. The operator reports that LLM agents performed most engine modifications under human-issued tasks, while the operator directed or approved the route and supervised review and deployment. Agents had no independent authority to initiate changes or deploy them. Machine recommendations also passed through human execution decisions. Section~\ref{sec:system} separates engineering, diagnosis, and trading roles.

During 2026, the account experienced poor production performance followed by a strong rebound. An adverse-regime-and-recovery narrative might fit that trajectory, but a forensic reconstruction could not establish it. Repository and configuration records document changes during the incident, while complete per-action deployment identity had not been persisted. The archive supports outcome and selected output descriptions; it does not establish a single frozen model entering and leaving a regime.

Our earlier Out-of-Money Reinforcement Learning (\oomrl) study emphasized \emph{evaluator independence}: externally imposed economic consequences can constrain agent-engineered evaluation \cite{liu2026oomrl}. Paper II examines \emph{diagnostic resolution}: those consequences do not identify their generating procedure or causal explanation. Independence here concerns an external constraint on internal evaluation narratives, not a complete correctness or alignment guarantee. The studies share some production observations, as disclosed in Section~\ref{sec:system}; the present question concerns the interpretation of feedback rather than replication of investment performance.

\subsection{From one non-stationarity to two}

The external environment can evolve while the software, configuration, artifacts, or adaptation rules that generate decisions also evolve. We use \emph{dual non-stationarity} descriptively to denote simultaneous evolution of the external environment and the evaluated decision procedure; we do not claim novelty for the phrase. Section~\ref{sec:framework} separates those changes formally.

This does not make prospective evaluation impossible. It requires precision about \emph{what object is being evaluated}. A frozen instance, a pre-specified adaptive procedure, and an ad-hoc development process support different longitudinal claims. Finance makes the distinction consequential: outcomes are stochastic, delayed, reference-dependent, and separated from recommendations by execution constraints; the historical market cannot literally be reset for debugging. Acceptance of the systems argument does not depend on believing that the strategy produces persistent alpha.

\subsection{Research questions}

\begin{description}[leftmargin=2.4cm,style=nextline]
\item[RQ1.] \textbf{What outcome- and output-level symptoms are supported by the available data under explicitly defined references, scopes, and horizons?}
\item[RQ2.] \textbf{Does the available evidence identify a single frozen production decision function or a unique technical root cause for the incident?}
\item[RQ3.] \textbf{What provenance and identity distinctions are required for scientifically valid longitudinal claims about frozen and adaptive agent-engineered systems?}
\end{description}

\subsection{Contributions}

The paper makes three contributions.

\begin{enumerate}[leftmargin=*]
\item \textbf{A longitudinal case with archive-bounded diagnostic findings.} We report one year of live account performance, a retrospectively identified incident, and conditional candidate-level and reference-relative symptoms. Style, cohort, and replay diagnostics constrain selected narratives. Development/configuration changes overlap the episode while action-level binding remains incomplete; the archive establishes neither a common frozen instance nor a unique technical cause.

\item \textbf{An evaluated-object identity and evidence framework.} We distinguish frozen instances, pre-specified adaptive procedures, and ad-hoc development, with per-execution state/input binding as an additional dimension. Compatible histories express archive-relative claim identifiability, illustrated through a loader example grounded in source history. An evidence hierarchy separates outcomes, reference-relative symptoms, subsystem attribution, and causal claims. Pre-specified adaptation can preserve a prospective record across state updates; exact factual identity does not supply counterfactual identification.

\item \textbf{A prospective production-binding protocol.} We propose grouped provenance and resolvable recommendation-to-runtime links, followed by separate links to executed actions and human substitutions. The design specifies future evidence for longitudinal diagnosis and preserves unresolved bindings explicitly. It has not been prospectively evaluated or shown to improve diagnosis in this case.
\end{enumerate}

\section{Conceptual Framework: Reality as Oracle, Not Debugger}\label{sec:framework}

\subsection{Restricted oracle and the outcome--diagnosis gap}
We use \emph{oracle} in a deliberately restricted sense: an external source of realized outcome evidence rather than a complete correctness specification. A financial loss is not a ground-truth software-fault label. Testing already distinguishes outcome acceptance from fault localization \cite{barr2015oracle,park2025flex,zeller2002delta}; production ML distinguishes detection, diagnosis, and reaction \cite{shankar2021observability}. Our concern is their interaction with the identity of a continually modified agent-engineered system.

\begin{figure}[htbp]
\centering
\includegraphics[width=\linewidth]{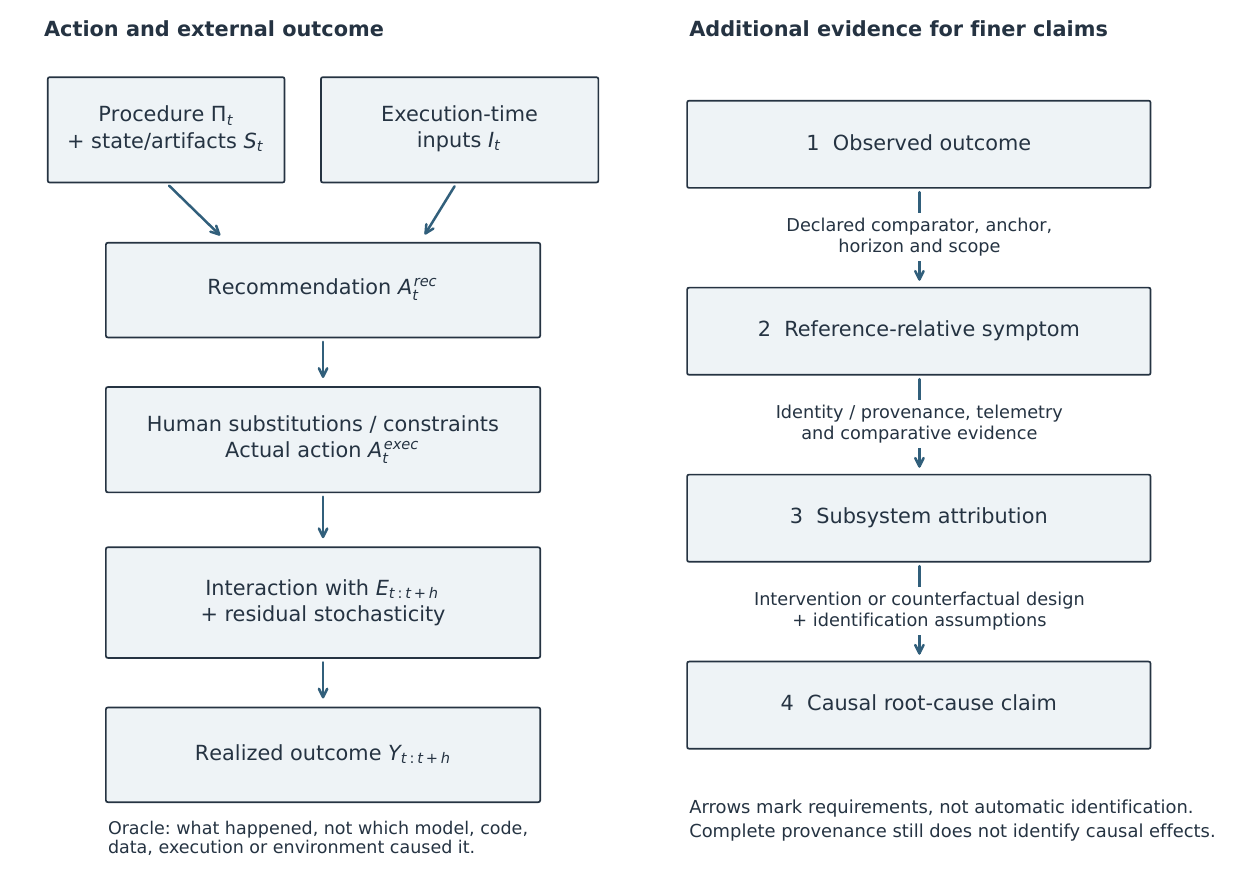}
\caption{Reality provides outcome evidence; diagnostic resolution requires additional structure. Recommendations pass through an execution path before producing account outcomes. The right-hand arrows mark evidentiary requirements, not sufficient conditions or an automatic route to causal identification. The levels characterize increasing claim granularity, rather than a mandatory diagnostic workflow.}
\label{fig:oracle-diagnosis}
\end{figure}

The \emph{outcome--diagnosis gap} is the difference between what an external outcome establishes and the strongest internal causal interpretation justified by available evidence. A poor outcome does not imply a software defect, a model defect, or a regime failure of a frozen model. Even a reference-relative symptom is conditional on comparator, timing, horizon, and scope. Figure~\ref{fig:oracle-diagnosis} summarizes the additional requirements for finer claims.

\FloatBarrier

\subsection{One action, outcome, and update model}
Let $\Pi_t$ be the declared decision procedure, $S_t$ its material internal state/artifacts, and $I_t$ execution-time inputs and context:
\begin{align}
 A_t &= \Pi_t(S_t,I_t), \label{eq:action}\\
 Y_{t:t+h} &= \mathcal{R}(A_{t:t+h},I_t,E_{t:t+h},\varepsilon_t), \label{eq:outcome}\\
 S_{t+1} &= U_{\Pi}(S_t,O_{\leq t}). \label{eq:update}
\end{align}
$E_{t:t+h}$ is the external environment over the outcome horizon and $\varepsilon_t$ residual stochasticity. The update relation applies when $U_{\Pi}$ is part of a procedure declared before the relevant future outcomes. $O_{\leq t}$ contains only observations available at that update, respecting delayed labels.

For account outcomes, $A_t$ means executed action under the full decision/execution procedure. Here machine recommendations $A_t^{\mathrm{rec}}$ precede human substitutions, cash allocation, and feasible fills $A_t^{\mathrm{exec}}$. These intermediate objects require separate bindings: candidate forward returns are not realized account P\&L. $A_{t:t+h}$ denotes the realized action path over the outcome horizon; $I_t$ includes the initial portfolio exposure relevant to that outcome; such operational inputs can vary even when fitted artifacts are frozen. Equation~\eqref{eq:outcome} is a schematic dependence model, not a fitted causal model. It therefore does not assign a multi-action portfolio return to one initiating recommendation.

\subsection{Evaluated-object identities}
\paragraph{Frozen instance.} The procedure $\Pi$ and relevant fitted/artifact state $S$ are fixed before evaluation; operational inputs, including portfolio inventory, can still vary. Material instance changes end the instance-specific record. A portfolio-return horizon involving later decisions under a changed instance cannot be assigned wholly to the earlier instance just because its initiating order preceded the change. A forecast issued under a bound instance can still be scored when its delayed label arrives after that instance retires; that forecast score is distinct from attributing the subsequent account path to it.

\paragraph{Pre-specified adaptive procedure.} The procedure and allowed update rule are fixed prospectively; state may evolve through scheduled retraining, online updates, or pre-declared model-selection rules. The scientific object is the procedure, not one fitted vector. A valid prequential record need not reset after each state update \cite{dawid1984prequential,gama2014drift}.

\paragraph{Ad-hoc development process.} Discretionary changes to procedure, update rules, model family, ensemble logic, or prediction pipeline outside declared adaptation channels create new procedure-level claim boundaries. An organization-level record can still be described, but cannot be relabeled as one frozen-instance or prospectively specified adaptive-procedure record.

\paragraph{Per-execution binding.} All these settings require realized artifact identities, data-as-of state, universe, overrides, and execution context. This is a binding dimension, not a fourth mutually exclusive procedure class. Not every parameter update or input change changes procedure identity.

\subsection{Evidence-compatible histories and archive-relative claim identifiability}\label{sec:compatible}
Let $\mathcal{E}$ be an evidence archive and $\mathcal{M}$ declared interpretation assumptions. Interpretation assumptions in $\mathcal{M}$ must be explicit and justified independently of the conclusion being evaluated; stipulating a disputed runtime binding is not archival evidence for that binding. $\mathcal{H}(\mathcal{E};\mathcal{M})$ denotes compatible deployment histories under those assumptions. Histories can specify procedure identities, artifacts, data snapshots/cutoffs, loaders, runtime configurations, overrides, recommendations, and execution. A claim functional $C$ is \emph{archive-identified} when this set is nonempty and
\[
 C(h_1)=C(h_2)\quad\text{for all }h_1,h_2\in\mathcal{H}(\mathcal{E};\mathcal{M}).
\]
An inconsistent archive with an empty set does not identify a claim. Identification means invariance across compatible histories, not confirmation: a claim can have an identified false value. This defines evidentiary resolution, not a new statistical estimator or causal method. Reliable additional evidence $\Delta$ can narrow the set:
\[
\mathcal{H}(\mathcal{E}\cup\Delta;\mathcal{M})\subseteq\mathcal{H}(\mathcal{E};\mathcal{M}),
\]
provided the new evidence is compatible and interpretation assumptions are fixed. Inclusion need not be strict. This is a conceptual archive-relative criterion, not a history-enumeration algorithm or an identification theorem proved for this case. We distinguish an established claim, a claim not established by the archive, and formal non-identification demonstrated by checked compatible-history witnesses. We do not enumerate or estimate the size of this case's history set.

Even $|\mathcal{H}(\mathcal{E};\mathcal{M})|=1$ for the factual deployment history does \emph{not} identify its causal effect. Exact identity does not supply counterfactual outcomes or eliminate environmental confounding \cite{hernanrobins2020}. Applying the criterion to a causal claim requires a model/history description rich enough to specify counterfactual assumptions; a unique factual trace alone is insufficient.

\paragraph{Illustrative example: resolving a referent without identifying an effect.}
For a recommendation with ID $a$, let $B_a(h)$ denote its procedure, artifact, and input binding in a history $h$. The source history records a March 2026 change from shared to model-mode-specific experiment lookup for predictions (Table~\ref{tab:history}). These are documented loader designs, not bindings to particular live recommendations.

Now consider a hypothetical archive retaining recommendation $a$, its output digest, associated actions and outcome, but no run-to-loader relation. Under fixed interpretation assumptions $\mathcal{M}$, suppose two histories match every retained record, with distinct bindings corresponding to the two loader designs:
\[
h_A,h_B\in\mathcal{H}(\mathcal{E};\mathcal{M}),\qquad
B_a(h_A)=b_A\ne b_B=B_a(h_B).
\]
Its factual binding is therefore not archive-identified. Suppose reliable additional evidence $\Delta$ recovers a run-captured relation from $a$ to $\mathrm{run}_B$ with resolvable, immutable code/configuration, artifact, and input snapshots. If the updated history set is nonempty and every member assigns $b_B$ to $a$, that referent becomes archive-identified; other parts of the history can remain unresolved.

This resolves which runtime produced the recommendation, not whether the other loader would have avoided a subsequent portfolio loss. A causal claim requires a defined intervention and comparative assumptions about later decisions, execution, and the environment. The historical change motivates the example; the recommendation, witness histories, and recovered binding are hypothetical, not reconstructed incident records. Section~\ref{sec:rq2} applies the distinction to the actual production archive.

\subsection{Evidence hierarchy}\label{sec:hierarchy}
\begin{description}[leftmargin=3.4cm,style=nextline]
\item[Level 1: observed outcome.] A directly recorded external outcome occurred.
\item[Level 2: reference-relative symptom.] Under a declared comparator, horizon, anchor, and scope, an output behaves poorly or changes direction.
\item[Level 3: subsystem attribution.] Identity, telemetry, and comparative evidence support association/localization to a subsystem or configuration class and distinguish reasonable alternatives. Association is not a causal effect.
\item[Level 4: causal root cause.] Evidence and identification assumptions support a mechanism, ideally through intervention, a natural experiment, or justified counterfactual design.
\end{description}
Reference design is needed for Level 2; finer attribution needs identity/provenance and comparison; Level 4 needs causal identification. These are requirements, not sufficiency guarantees. Provenance is necessary for fine-grained attribution, but is not causal magic.

\section{System and Evidence}\label{sec:system}

\subsection{Production setting}

\sys{} is a live quantitative equity-selection and portfolio-management system. During the focal period it operated on a CSI300-related stock subset and typically generated recommendations on a weekly-like batch schedule; exact historical batch dates, rather than a verbal ``weekly'' label, define the realized schedule.

\paragraph{Engineering and diagnostic roles.} The operator describes a human-directed workflow: LLM agents modified code and configuration, drafted plans, and conducted requested experiments. Most engine implementation changes were agent-assisted, according to the operator; no numerical authorship census is claimed. The operator guided or confirmed the direction and supervised review and deployment. Agents could act within assigned tasks, but had no authority to independently initiate changes or production deployment. This human-directed engineering and diagnosis workflow is the operational meaning of \emph{agent-engineered} in this case.

Agent-assisted engineering makes human-issued tasks, tool actions, proposed code/configuration changes, and acceptance decisions relevant parts of engineering lineage. A prospective record would connect those steps to the artifacts actually used at runtime, then link recommendations separately to human substitutions and executed actions. These relations describe evidence to preserve, rather than a chain fully recovered from the historical archive. Engineering prompts need not be per-order LLM inference inputs.

Contemporaneous configuration notes also document an LLM Critic producing diagnostic proposals. The 27 May note records prompt rules requiring suggested changes to be distinguished from executed ones. A 26 June Critic trace retained the sliding-window rule, yet proposed changing the nominal \texttt{fit\_end\_time} field that did not control that boundary under the recorded slide mode. The 28 June configuration note records that the proposal was skipped. Inspected historical code derives the slide-mode boundary from the anchor and validation/test windows. These records illustrate the distinction between a generated proposal, its runtime semantics, and its disposition, rather than identifying the cause of portfolio loss. The note's root-cause narrative is not adopted here; Table~\ref{tab:history} identifies the historical changes and note sources, and the trace identity is retained in the local research audit.

The production path was not fully automatic. Historical operation included manual judgment about opening tradability, candidate substitution, round-lot execution, cash allocation, and other practical execution decisions. Recommendation inventory distinguishes primary selections, alternates, and the full candidate set. These objects must not be conflated with actual purchases.

Recommendations were generated from information available after the preceding market close; the date in an order filename denotes intended execution, not the availability of that day's prices. At the intended day's open, the operator sold designated positions in full and allocated remaining cash approximately equally across selected buys. Buys were in integer lots of 100 shares. A candidate inventory approximately three times the primary selection count supported substitutions when a stock could not be bought. Opening tradability, including suspension and limit-state judgments, was determined manually; these judgments were unavailable when recommendations were generated. Suggested prices and quantities were references rather than executable orders. Unused cash remained in the account. Actual account returns include recorded settlement costs; configured backtest fee rates are used only in the stated approximate simulations.

\subsection{Analysis-wide annual window}

The annual interval was fixed for the present retrospective analysis at the 30 September 2026 evidence freeze; it was not preregistered before the outcomes occurred. The analysis-wide window is
\[
\text{2025-10-01 through 2026-09-30},
\]
using the 30 September 2025 close as the wealth baseline and the first return observation on 9 October, for 241 daily return observations.

The descriptive segments are:
\begin{itemize}[leftmargin=*]
\item initial period (\emph{Initial} in tables and figures): through 28 February 2026, $n=94$;
\item incident-background period (\emph{Mar--Jun}): 1 March--30 June 2026, $n=82$;
\item later period (\emph{Jul--Sep}): 1 July--30 September 2026, $n=65$.
\end{itemize}
The complete annual interval is labeled \emph{Full year}.
The April--June narrative is an incident subwindow and is not identical to the March--June statistical segment. Actual return dates are 9 October--27 February, 2 March--30 June, and 1 July--30 September; preceding-close baselines are 30 September 2025, 27 February 2026, and 30 June 2026, respectively. The initial 94-day segment overlaps the production phase reported in Paper I; it is not an independent replication. Paper I does not establish persistent alpha, a causal module effect, or a frozen-model OOS record.

\subsection{Account returns and cash flows}

Let $V_t$ denote recorded closing account value, including equity valuation and signed cash, and let $CF_t$ denote external cash flow, positive for inflows. Following the production analyzer's beginning-of-day cash-flow convention,
\[
r_t=\frac{V_t-V_{t-1}-CF_t}{V_{t-1}+CF_t},\qquad
W_t=\prod_{u\leq t}(1+r_u).
\]
Missing cash-flow fields are treated as zero and adjusted capital must be positive. Small recorded negative cash balances are retained; no additional financing cost or fictitious funding is inserted. The operator identifies the 27 February sale as a partial discretionary cash-out for purposes outside the quantitative strategy. It did not represent a complete portfolio liquidation and was unrelated to the research cutoff. This annotation does not change automatic trade classification and does not imply an external withdrawal beyond the cash-flow ledger. This daily convention does not reconstruct intraday funding timing.

\subsection{Reference series}

The paper uses the CSI300 index as one broad benchmark and an idealized equal-weight reference over the CSI300-related stock subset as an additional cross-sectional context. The latter is \emph{not} described as an executable trading strategy. The subset follows system-specific eligibility rules and is not the full CSI300 universe. The reference uses prior-trading-day membership, recorded delisting exclusions, and daily ideal gross rebalancing. It carries forward the last available positive adjusted close for missing prices, without backward filling from future observations. A carried suspension price contributes zero price return until a new quote is available, when the price change is recognized. This is a valuation convention, not an assertion of tradability. We therefore refer to it as an \emph{approximate equal-weight reference}. Annual reference return is $-5.1523$\%; all 241 return dates are covered, with carried endpoints on 66 dates. CSI300 benchmark returns use Qlib adjusted closes; comparison with the account ledger confirms agreement within recorded rounding and provider precision.

Candidate-relative returns are available in a signal-close-matched construction: for each candidate, the approximate equal-weight wealth change is measured over the same signal-close-to-horizon interval. No corresponding execution-open-matched equal-weight series has been constructed. Signal-close-relative and execution-open analyses are therefore not interchangeable.

\subsection{Candidate horizons}

A recommendation date is the intended execution date. The \emph{signal-close} anchor is the previous available close; the \emph{execution-open} anchor is the intended execution-day open. Under the stated horizon convention, $h1$ ends at the execution-day close, while $h20$ ends at the close of the twentieth trading day starting from the execution day. We evaluate $h\in\{1,3,5,10,15,20\}$, with terminal trading-day index equal to execution index plus $h-1$. The previous trading-day close is a measurement proxy for signal timing, not a recovered per-action data-as-of timestamp. Candidate prices are observed adjusted endpoints without stale-price filling; missing prices are distinguished from a target after the cutoff. Of 7,596 requested candidate/anchor/horizon labels, 7,366 are observed and 230 are right-censored at 30 September. These are label counts across both anchors and all horizons, not distinct stocks. No later outcomes are used. Candidate-relative excess is the candidate return minus the compounded daily approximate reference over the same signal-close interval; beating the reference means strictly positive excess.

\subsection{Statistical convention}

The main display uses a single-factor daily excess-return regression,
\[
r_t-r_f/252=a+\beta(b_t-r_f/252)+\epsilon_t,
\]
where $b_t$ is CSI300 daily return and $r_f=0.0135$ per year. Annualized alpha is $252a$. We adopt HAC covariance with lag 3, the Bartlett kernel, \texttt{use\_t=True}, and small-sample correction as the main convention for this retrospective analysis; it was not a prospectively registered test. OLS, HAC(5)/HAC(10), and $r_f=0$ are retained as sensitivity analyses. Table~\ref{tab:covariance-sensitivity} displays the existing market-only OLS/HAC grid. Lag 3 is a fixed presentation convention for this retrospective analysis; no record establishes a theoretically optimal or prospectively selected lag. HAC addresses heteroskedasticity and finite-lag serial correlation of fitted daily regression residuals and changes covariance, not coefficients \cite{neweywest1987}. It does not correct retrospective window selection or multiple testing, nor supply valid inference for the separate repeated-candidate and overlapping-horizon summaries.

For $n$ daily observations, total return is $R=\prod_t(1+r_t)-1$, CAGR$_{252}=(1+R)^{252/n}-1$, volatility is $\sqrt{252}\,s_r$, and Sharpe is $(252\bar r-r_f)/(\sqrt{252}\,s_r)$, using sample standard deviation $s_r$. CAGR$_{252}$ is trading-day annualization, not elapsed-calendar-time CAGR. Drawdown includes initial wealth 1.

Multi-factor sensitivity adds lagged liquidity, momentum, and volatility proxies to market excess return. These are engine-inspired proxies, not canonical Fama--French, Barra, or industry-neutral factors. Detailed construction and legacy-definition sensitivity are retained in Appendix~\ref{app:factor-methods}; specifications are not selected for significance.

\subsection{Recorded evolution and temporal ambiguity}
\begin{figure}[htbp]
\centering
\includegraphics[width=\linewidth]{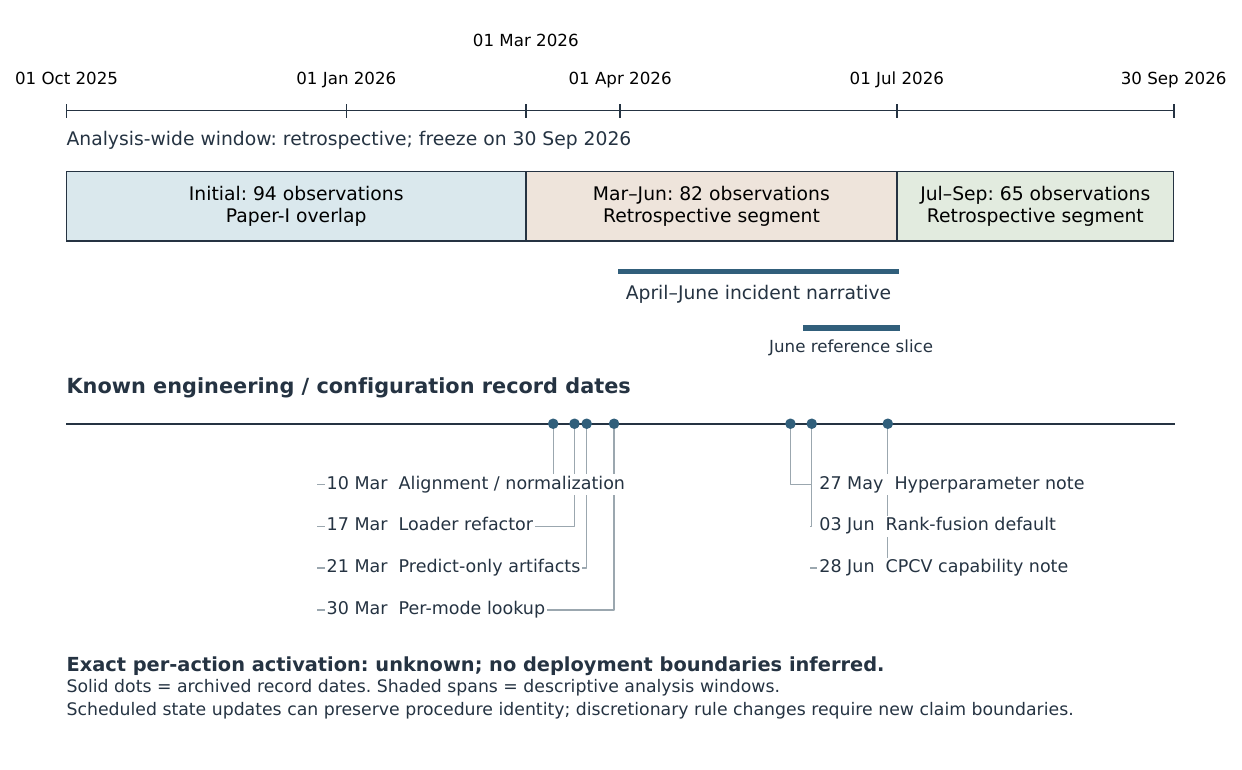}
\caption{Retrospective spans and known engineering/configuration record dates. Counts are trading-day return observations within calendar spans. The March--June statistical segment differs from the April--June incident narrative. Solid dots identify record dates only; exact per-action activation is unknown. No deployment epoch is inferred from a commit or note date.}
\label{fig:production-timeline}
\end{figure}

Figure~\ref{fig:production-timeline} places analysis episodes beside inspected engineering/configuration record dates. Their overlap motivates the identity question; it does not establish production activation boundaries. Supplementary Dataset S1 supplies the daily account/reference paths and pseudonymous closing holdings separately from the manuscript (Appendix~\ref{app:availability}). Table~\ref{tab:history} retains source identifiers. The annual window and Paper-I overlap are analytical spans, not evidence of one instance throughout.

\FloatBarrier

\section{Annual Context and Incident}\label{sec:annual}

Table~\ref{tab:periods} reports the fixed-for-analysis annual result and the three retrospectively described segments using the stated HAC(3) convention.

\begin{table}[htbp]
\centering
\caption{Production account and CSI300 outcomes. Segment boundaries are retrospective descriptions, not pre-registered hypothesis tests. Alpha $p$-values use HAC(3), \texttt{use\_t=True}, with small-sample correction.}
\label{tab:periods}
\small
\resizebox{\linewidth}{!}{%
\begin{tabular}{lrrrrrrr}
\toprule
Period & $n$ & Account & Index & CAGR$_{252}$ & Sharpe & Ann. $\alpha$ & HAC3 $p$ \\
\midrule
Full year & 241 & +11.1391\% & -6.0997\% & +11.6762\% & 0.6726 & +14.4911\% & 0.27526 \\
Initial & 94 & +13.1731\% & +1.5075\% & +39.3406\% & 2.0598 & +30.0744\% & 0.03339 \\
Mar--Jun & 82 & -14.7887\% & +5.7058\% & -38.8486\% & -2.8108 & -58.1058\% & 0.01778 \\
Jul--Sep & 65 & +15.2461\% & -12.4876\% & +73.3488\% & 3.6873 & +76.5417\% & 0.00230 \\
\bottomrule
\end{tabular}
}
\end{table}

The annual record is positive relative to the index but does not establish a statistically significant alpha under the chosen annual single-factor specification: annualized alpha is 14.49\% with HAC(3) $p=0.275$. The corresponding multi-factor intercept is 13.52\% with HAC(3) $p=0.226$.

For the 94-day overlap with Paper I's Phase 3, the OLS estimate reproduces its reported rounded annualized alpha of 30.07\% and $p=0.094$. This is a comparison of reported values, not a new verification of Paper I's original inputs. The HAC(3) presentation changes the covariance convention: the Initial segment's 5\% significance classification is sensitive to that convention (Table~\ref{tab:covariance-sensitivity}).

The segmented estimates differ sharply: the initial and later intercept estimates are positive, and the March--June estimate is negative under each stated specification. These signs do not define three pre-specified statistical regimes. The windows were selected after observing the production trajectory; the associated $p$-values describe uncertainty conditional on that sample and do not repair the selection step. Multi-factor HAC(3) $p$-values for the initial and later segments are 0.022 and 0.0079, respectively.

For the selected March--June interval, the market-only annualized intercept is $-58.1058$\%. Its HAC(3) $p$-value is 0.01778. Adding the three lagged style proxies on the same 82 observations changes the intercept to $-25.3155$\% ($p=0.22919$), with a 95\% interval of $[-66.9051\%,+16.2741\%]$. Both the intercept's absolute magnitude and its standard error decrease. Thus intercept inference is sensitive to the explanatory variables, while the observed account loss remains $-14.7887$\%. The two significance classifications are not a test of the difference between estimates \cite{gelman2006significance}. The change is compatible with exposure-related alternative explanations; it does not identify a market mechanism, prove the absence of model failure, or correct retrospective window selection.

The annual estimate is imprecise while selected subperiod estimates differ sharply. These observations concern the evolving account-level process and do not identify a single stationary model.

\begin{figure}[htbp]
\centering
\includegraphics[width=0.96\linewidth]{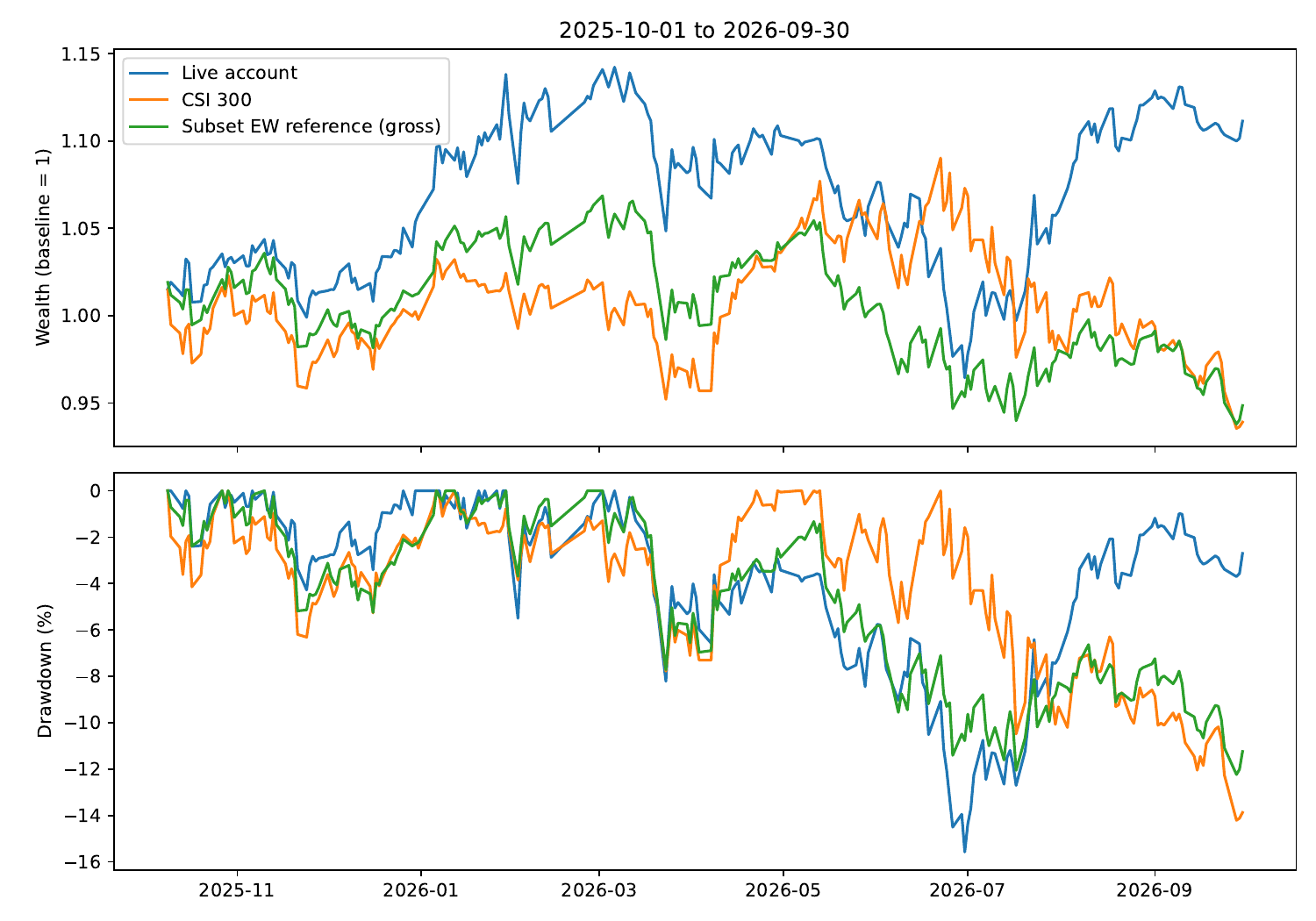}
\caption{Annual account, CSI300, and approximate subset EW wealth and drawdown. Account returns use recorded execution and beginning-of-day cash-flow adjustment; EW is a gross ideal reference with carried suspension prices.}
\label{fig:annual_nav_drawdown}
\end{figure}

\subsection{Monthly context}

Table~\ref{tab:monthly} reports all twelve months using contemporaneous close-to-close return dates. March account return was $-4.2922$\%, compared with CSI300 $-5.5321$\% and approximate EW $-6.0845$\%; an absolute March loss alone does not establish reference-relative degradation. August account return was +6.1394\%, followed by $-1.1940$\% in September.

The complete twelve-month numerical table is retained in Appendix~\ref{app:monthly}.

\begin{figure}[htbp]
\centering
\includegraphics[width=0.96\linewidth]{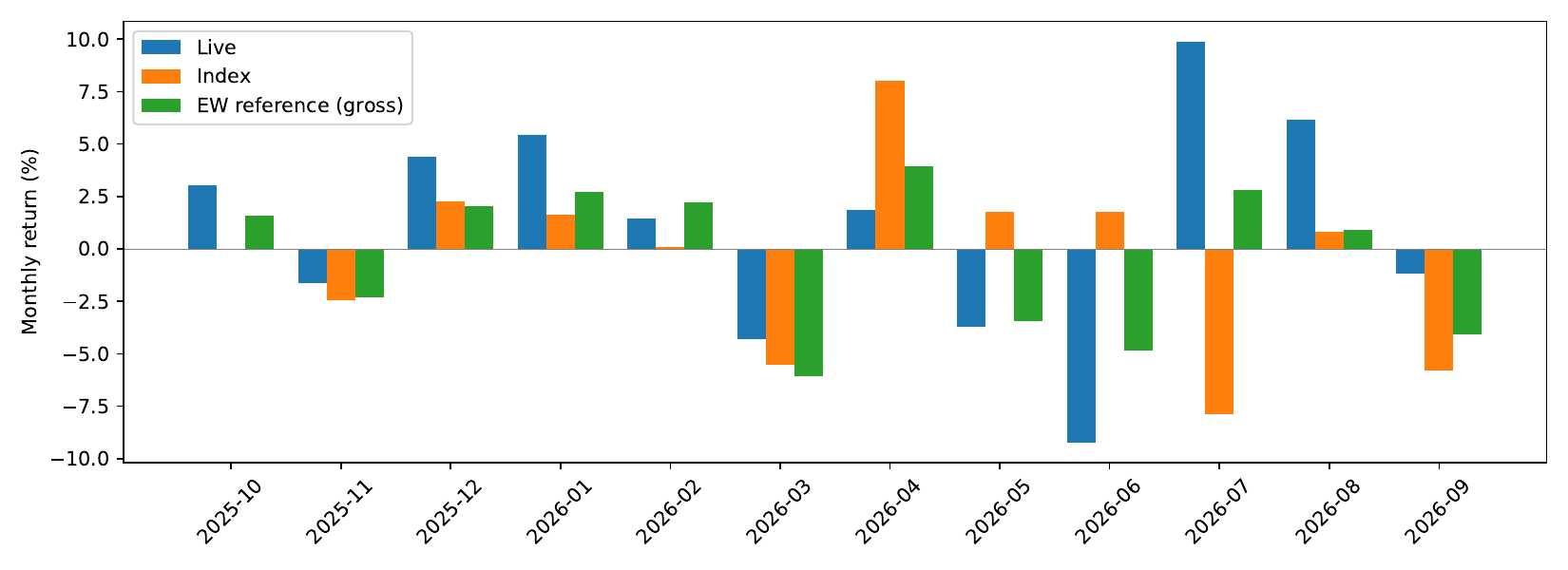}
\caption{Monthly account, CSI300, and approximate gross EW returns across the complete annual period, without a one-day-forward benchmark shift.}
\label{fig:monthly_returns}
\end{figure}

\section{RQ1: What Production Symptoms Are Actually Supported?}\label{sec:rq1}

RQ1 examines account outcomes and output-level symptoms relative to declared anchors, candidate scopes, and references. It does not presuppose that every layer deteriorated from a uniformly healthy baseline.

\subsection{Candidate classes are not interchangeable}

The production archive distinguishes primary recommendations, alternates, and all candidate occurrences. ``All candidates'' are counted by recommendation occurrence and are not independent-stock samples or actual purchases. Actual execution can include manually selected alternates. Consequently, a candidate analysis cannot be substituted for account P\&L, and a primary-candidate result cannot be silently generalized to all candidates.

The scope contrast already appears in the initial period. At T+10 relative to approximate EW, January and February primary candidates have negative median excess returns ($-2.1662$ pp, $n=13$, and $-4.0636$ pp, $n=10$), while all candidates have positive medians (+1.6262 pp, $n=39$, and +0.8827 pp, $n=30$). These occurrence summaries do not establish a uniformly healthy early layer or a statistically tested cross-period deterioration.

\subsection{T+10 absolute and relative symptoms}

For April and June, the signal-close-matched T+10 construction shows negative candidate-relative symptoms against the approximate equal-weight reference.

\begin{table}[htbp]
\centering
\caption{Selected T+10 relative symptoms using the signal-close anchor and approximate gross equal-weight reference. Values are candidate-occurrence statistics, not actual trade P\&L.}
\label{tab:relative}
\small
\begin{tabular}{llrrr}
\toprule
Month & Candidate scope & $n$ & Median excess & Beat fraction \\
\midrule
April & all candidates & 35 & -2.1250 pp & 28.57\% \\
April & primary & 12 & -2.5038 pp & 25.00\% \\
June & all candidates & 45 & -1.9593 pp & 28.89\% \\
June & primary & 15 & -1.9593 pp & 33.33\% \\
\bottomrule
\end{tabular}
\end{table}

Primary occurrences are a subset of all candidates. In June, the same primary occurrence is the 23rd of 45 all-candidate observations and the 8th of 15 primary observations when ordered by excess return, so the two unrounded medians coincide. The beat fractions and other distributional summaries differ; the identical medians do not make the two scopes interchangeable.

May demonstrates why absolute and relative metrics must remain separate. At T+10, all candidates have an absolute median return of $-1.7557$\% but a relative median of +0.4430 pp; primary candidates show $-0.8288$\% absolute and +1.0277 pp relative. A poor absolute outcome can therefore coexist with a positive reference-relative outcome.

The appropriate RQ1 claim is limited: \emph{for specific horizons and candidate scopes, April and June contain negative relative selection symptoms against this approximate reference}. All four selected cells are fully observed. The analysis does not prove environment-independent causal failure.

\begin{figure}[htbp]
\centering
\includegraphics[width=0.96\linewidth]{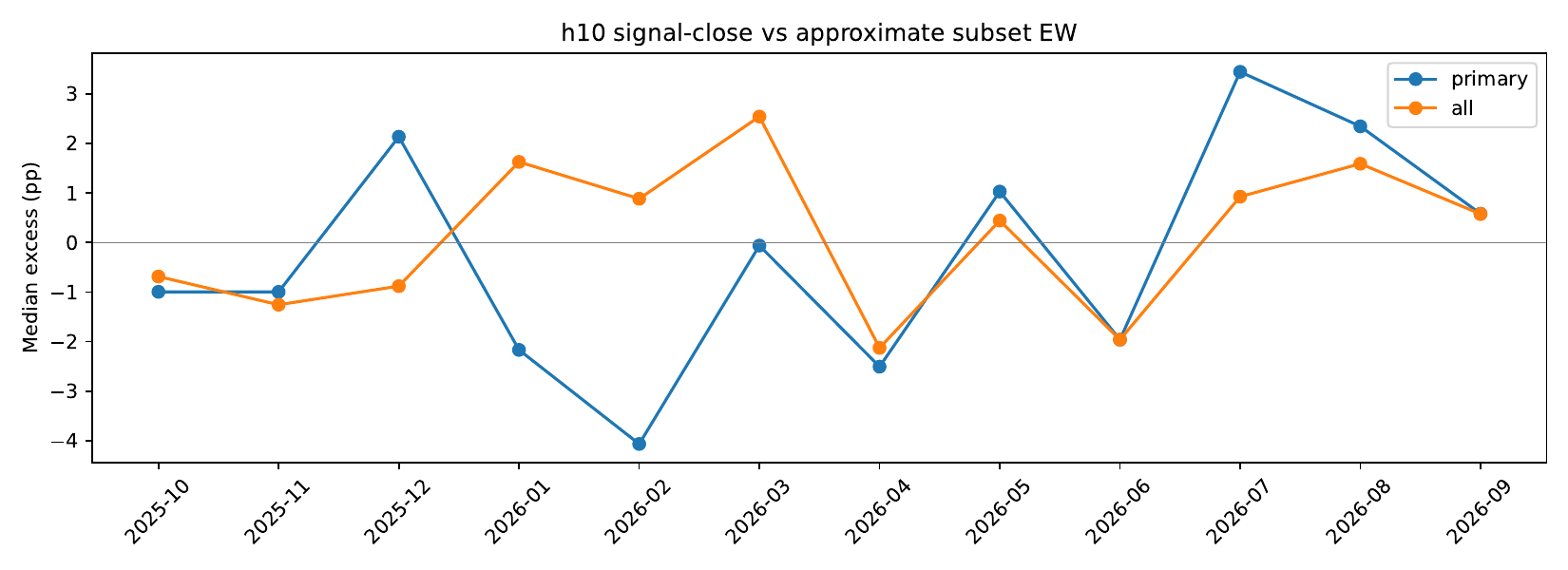}
\caption{T+10 candidate-occurrence excess returns against approximate gross EW, using signal-close matching and separate primary/all scopes. Observed counts for the selected April/June cells appear in Table~\ref{tab:relative}; all observed/requested counts are retained in the local research archive.}
\label{fig:candidate_relative_h10}
\end{figure}

\subsection{Diagnostic narratives tested against the archive}\label{sec:diagnostic-narratives}
Secondary diagnostics answer specific questions about retrospective explanations. Their value is to constrain an explanation's scope and specify the next evidence needed, rather than to turn every symptom into a root cause. Appendix~\ref{app:secondary} retains the definitions and numerical details.

\paragraph{Style exposure.} One retrospective explanation attributes the incident to changed market exposure. Section~\ref{sec:annual} reports the market-only and multi-proxy estimates on the same 82 observations. Their specification sensitivity keeps exposure-related explanations open; actual exposures would need independent evidence before a mechanism could be established.

\paragraph{Cohort accounting.} A second explanation singles out May entries as the main source of June's loss. The fixed May-29 basket and the account-linked decomposition measure different objects. In the latter, May-labeled instruments contribute $-2.84050$ pp, while instruments outside that fixed cohort contribute $-5.06181$ pp, the largest negative group. This constrains the aggregate accounting claim and redirects attention beyond the May group. The outside group is not all June acquisitions, and its size and contribution do not identify a causal stock, trade, or subsystem.

\paragraph{Avoided trading.} A third explanation assumes that retaining earlier holdings would have prevented the loss. From the 2 March close, fixed-unit freeze returns $-16.6910$\% versus matched production $-15.4779$\%; an April snapshot produces a different ordering. Drop0/1/3/6/9 replay is non-monotonic under the stated approximate execution rules. These results constrain a simple ``less trading would have solved it'' ranking within this design. The snapshot and execution limitations are specified in Appendix~\ref{app:secondary}; this comparison does not identify an optimal live policy or failure onset.

Other measurements retain their own scope: recurrence is not artifact freshness, Sell-side effects remain open, and mixed Swap and horizon results do not support a universal replacement or Buy-signal failure.

\subsection{Approximate random-ranking reference}

The reference uses the engine's pure numerical TopK Dropout random ranking with $K=22$, 5,000 continuous annual paths per profile (Drop0/3/22), seed 20261001, and five-trading-day constituent updates. June is sliced from those paths, not restarted at actual June holdings. Ideal daily equal weighting, carried prices, and proportional turnover fees omit round lots, minimum fees, opening fills, and human substitutions; suspended stocks may be selected despite being non-executable. Appendix~\ref{app:random-methods} retains full mechanics. This is an approximate comparative reference, not a matched production null.

For June under the Drop3 reference:
\begin{itemize}[leftmargin=*]
\item production account return: $-9.2297$\%;
\item simulation P5: $-9.3076$\%;
\item simulation median: $-5.1439$\%;
\item simulation P95: $-0.1505$\%;
\item lower fraction at or below production: 272/5,000 = 5.44\%;
\item 94.56\% of simulated paths are above production (with no ties in this sample).
\end{itemize}

This is a \emph{selected-period descriptive position in an approximate reference distribution}, not a root-cause $p$-value and not a strictly matched test of the production process.

The annual approximate Drop3 random reference provides an important counterweight: 51/5,000 paths return at least as much as the observed production year, an empirical upper fraction of 1.02\%. Thus the observed account process can look unusually poor in a retrospectively selected month and unusually strong over the full year. The contrast provides context for interpreting the selected month. These summaries also answer different reference questions from regression inference: the annual market-only HAC(3) $p=0.27526$ concerns a zero intercept under the specified factor model, whereas 51/5,000 locates the account path within one mechanically defined uninformed-selection distribution. Its empirical tail fraction is not a competing calibrated alpha test, skill probability, or causal attribution test.

\begin{figure}[htbp]
\centering
\includegraphics[width=0.96\linewidth]{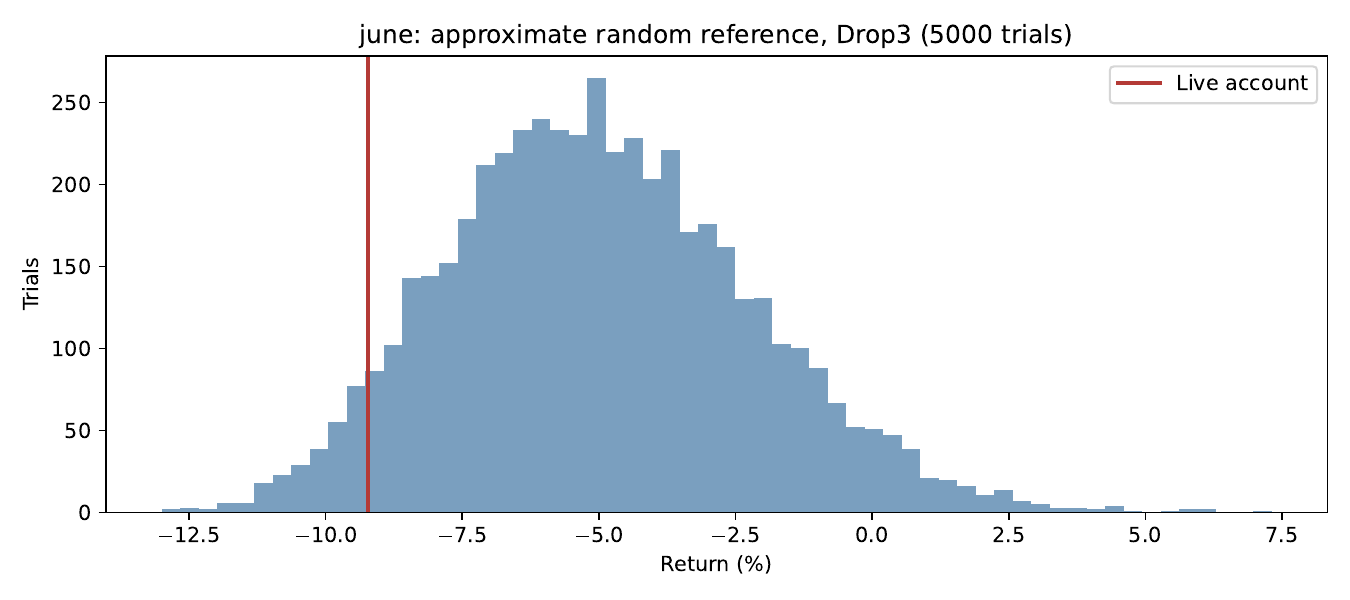}
\caption{June Drop3 approximate random-ranking reference: actual return $-9.2297$\%, with 272 of 5,000 paths at or below it. June is sliced from annual simulations and was selected retrospectively.}
\label{fig:random_june}
\end{figure}

\begin{figure}[htbp]
\centering
\includegraphics[width=0.96\linewidth]{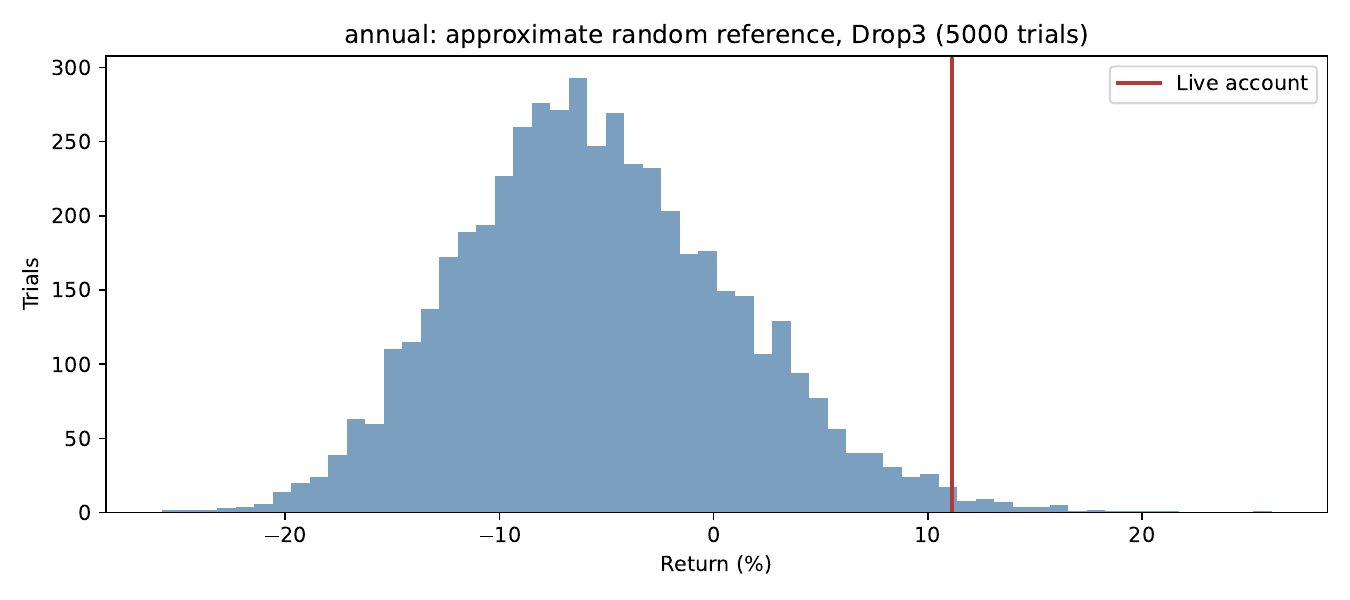}
\caption{Annual Drop3 approximate reference: 51 of 5,000 paths equal or exceed observed +11.1391\% account return. Execution and weight approximations are the same as for the June reference.}
\label{fig:random_annual}
\end{figure}

\subsection{RQ1 answer}

\begin{quote}
\textbf{RQ1 finding.} The data support limited, explicitly conditioned output-level symptoms rather than a universal ``Buy-signal failure.'' April and June show negative T+10 candidate-relative outcomes under the signal-close, approximate equal-weight reference; 272 of 5,000 approximate random-ranking paths (5.44\%) have June returns at or below the observed account return under the stated reference mechanics. Other horizons, candidate layers, Sell observations, and cohort decompositions are mixed. Secondary diagnostics constrain the scope of specific narratives and the next questions to investigate; they do not establish a cause independent of market conditions or production-stack changes.
\end{quote}

\section{RQ2: What Can Be Attributed to the Evolving Production Stack?}\label{sec:rq2}

\subsection{Repository/configuration history is not deployment identity}

Repository changes and production configuration notes document evolution in prediction infrastructure, normalization, model hyperparameters, and training capabilities (Table~\ref{tab:history}). They establish that changes existed in the development/configuration history; demonstrated use in a live recommendation requires action-level runtime binding. This history describes an evolving stack, not a complete deployment timeline.

\subsection{Prediction-infrastructure changes}

Inspected diffs record changes to prediction loading, per-mode recorder lookup, predict-only artifacts, and ensemble normalization. Configuration notes record hyperparameter revisions and the introduction of CPCV training capability. These records motivate engineering explanations to examine, without establishing a particular contaminated order.

\begin{table}[htbp]
\centering
\caption{Inspected historical evidence of development/configuration changes. Commit dates and note dates are record dates, not established per-action activation dates. Model identities and private configuration contents are omitted.}
\label{tab:history}
\small
\begin{tabularx}{\linewidth}{p{2.2cm}p{2.6cm}X}
\toprule
Record date & Source identifier & Recorded change \\
\midrule
10 Mar 2026 & Git \texttt{59e2977} & Ensemble prediction/backtest alignment and normalization changes. \\
17 Mar 2026 & Git \texttt{7af9532} & Prediction-loader refactoring toward shared Qlib Recorder infrastructure. \\
21 Mar 2026 & Git \texttt{2194d1d} & Predict-only source-artifact loading and model persistence changes. \\
30 Mar 2026 & Git \texttt{2b5b23a} & Per-mode experiment lookup for recorders. \\
27 May 2026 & Configuration note & Hyperparameter revisions and reported production retraining. \\
3 Jun 2026 & Git \texttt{c2fa67a} & Percentile-rank normalization introduced as the default fusion option. \\
28 Jun 2026 & Configuration note & CPCV training capability and configuration documented. \\
\bottomrule
\end{tabularx}
\end{table}

The private note sources are \texttt{CHANGELOG\_2026-05-27.md} and \texttt{CHANGELOG\_2026-06-28.md} in the production workspace; the latter's root-cause narrative is not adopted as an identified cause here. Source hashes and inspected commit identities are retained in the local research audit. The limited runtime inventory contains 50 run metadata records (49 declared successful, one failed) and six cycle records (one declared sealed complete, five partial). These declarations do not independently verify all underlying actions, and this is not complete annual deployment binding.

\subsection{What is established and what remains open}

The strongest evidence statements are:
\begin{itemize}[leftmargin=*]
\item the annual and incident-period account outcomes are observed;
\item specific candidate-relative T+10 symptoms are observed under a declared approximate reference;
\item the development/configuration stack changed during the broad incident period;
\item complete per-action production identity is not established by the inspected archive.
\end{itemize}

Table~\ref{tab:claim-main} separates these observations from claims about a common frozen instance or a unique technical cause that remain unestablished. We describe the episode as \emph{conditional production symptoms under an evolving stack}. The later rebound is another outcome of the evolving account process; the archive does not bind it to the same frozen instance or identify a repair mechanism.

\subsection{Evidence-to-claim boundary}
Table~\ref{tab:claim-main} connects the case to Section~\ref{sec:hierarchy}. ``Not established'' denotes an archive-bounded lack of support, not a false claim or a checked proof of formal non-identification. Incomplete binding does not prove that an explanation never occurred. Appendix~\ref{app:evidence-matrix} retains the extended matrix.

\begin{table}[htbp]
\centering
\caption{Claim granularity supported by the archive. Levels describe proposed claims, not automatic certification.}
\label{tab:claim-main}
\small
\begin{tabularx}{\linewidth}{p{0.23\linewidth}p{0.12\linewidth}p{0.17\linewidth}X}
\toprule
Claim & Level & Status & Basis / missing requirement \\
\midrule
Annual account outcome & 1 & Observed & Recorded returns; no persistent-alpha inference. \\
April/June T+10 weakness & 2 & Conditional & Signal-close, scopes, approximate EW; not a market-independent cause. \\
June poor under random ranking & 2 & Descriptive & 272/5,000 lower paths; not a matched null. \\
March loss proves relative failure & 2 & Unsupported & Account exceeded both cited March references. \\
Universal Buy failure; optimal Drop3 & 2 & Unsupported & Mixed horizons/scopes; non-monotonic replay. \\
One frozen instance spans incident & Identity condition & Not established & Action binding $B_a(h)$ is unresolved across the archive. \\
Recorder/ensemble change caused loss & 4 & Not established & Factual identity and a justified causal comparison are both missing. \\
Later rebound is same-model recovery & 4 & Not established & Common frozen identity and recovery mechanism not established. \\
Provenance alone proves causality & General rule & False & Factual history is not counterfactual identification. \\
\bottomrule
\end{tabularx}
\end{table}

The example in Section~\ref{sec:compatible} illustrates how runtime bindings can resolve a claim's referent while leaving its causal interpretation open. In the actual case, complete bindings for the historical actions examined are not established. We have not constructed checked compatible-history witnesses proving formal non-identification of a historical recommendation's binding. Missing records alone do not show that both loader versions ran or produced different outcomes.

Resolved factual bindings would still require a defined intervention, comparative evidence, and counterfactual assumptions for a causal claim. Returns and declared labels support observed outcomes and conditional symptoms; approximate replay supplies a scoped reference rather than a controlled causal comparison. This is the distinction between identifying the evaluated object and explaining its result.

\subsection{RQ2 answer}

\begin{quote}
\textbf{RQ2 finding.} The available archive does not support a clean frozen-model regime-failure narrative or a unique technical root cause. It supports conditional production symptoms during an evolving development/configuration process whose action-level runtime bindings remain incomplete.
\end{quote}

\section{RQ3: Provenance-Bounded Evaluation}\label{sec:rq3}

\subsection{Choose and bind the scientific object}
Section~\ref{sec:framework} defines the identities. Frozen-instance claims require prospective fixation and per-action binding. Pre-specified adaptive procedures can continue a prequential record across legitimate state updates; information timing and allowed adaptations must be declared before outcomes. Discretionary revisions outside those channels create new procedure-level boundaries.

The annual record describes the evolving account process, not a demonstrated single frozen instance or fully pre-specified adaptive procedure. Action binding should reduce ambiguity about the referent before higher-level causal interpretations are attempted.

\subsection{Grouped per-action provenance manifest}
Table~\ref{tab:manifest} proposes a grouped provenance design derived from the case. It is not a new standard or evidence of historical compliance. PROV-DM supplies entities, activities, agents, and derivations; PROV-AGENT extends tracing to agent interactions \cite{w3c2013prov,souza2025provagent}. A compatible schema does not itself establish archive completeness.

\begin{table}[htbp]
\centering
\caption{Proposed per-action groups; full-year historical compliance is not established. Identifiers bind immutable content rather than mutable ``latest'' pointers.}
\label{tab:manifest}
\small
\begin{tabularx}{\linewidth}{p{0.21\linewidth}X}
\toprule
Group & Required fields / bindings \\
\midrule
Information timing & Signal available-after, market as-of, planned execution date, calendar identity/version. \\
Procedure identity & Procedure ID, source commit, configuration hash, prediction-loader version. \\
Model state & Ensemble identity/members, artifact IDs, training mode/cutoff, recorder IDs. \\
Decision context & Universe policy/snapshot hash, primary/alternate candidates, base and ensemble prediction hashes, final rankings. \\
Human/action path & Suggested orders, overrides/substitutions, actual orders/fills, observed constraints, fee model and actual fees with distinct roles. \\
\bottomrule
\end{tabularx}
\end{table}
Hashes of persisted inputs and suggestions strengthen integrity. Actual feasibility observations must be separated from replay limit approximations. Unavailable fields should be explicit, not inferred from later data. Manifest completeness and causal identification remain separate requirements.

\paragraph{First implementation priority: action-to-runtime binding.}\label{sec:minimal-binding}
For this case, a practical starting point is a resolvable relation: recommendation ID to run/instance ID to immutable procedure, configuration, artifact, and input references. Persist generation time and an output digest alongside content-backed snapshots captured during that run. A commit or an opaque hash cannot recover unrecorded runtime state. Separate the engineering-task lineage described in Section~\ref{sec:system} from generation-time prompt or agent-trace identities when those components actually affect the output.

Link executed-action/fill IDs separately to recommendations, substitutions, or an explicit manual/unlinked status. This preserves the distinction between a machine proposal and the action whose settlement affects the account. Missing bindings must remain unavailable rather than being filled retrospectively from mutable latest pointers. These links make the referent resolvable; claim level still depends on reference design, telemetry/comparison, and causal identification (Section~\ref{sec:hierarchy}).

\FloatBarrier
\par\addvspace{\medskipamount}
\noindent
\begin{minipage}[t]{\linewidth}
\subsection{Prospective protocol for frozen instances}
\begin{enumerate}[leftmargin=*]
\item Freeze relevant code, configuration, and artifacts before evaluation.
\item Bind each action to the instance and execution-time inputs; separate recommendations and executed actions.
\item End the instance-specific record at material change and explicitly handle horizons spanning that change.
\end{enumerate}
\end{minipage}

\subsection{Prospective protocol for adaptive procedures}
\begin{enumerate}[leftmargin=*]
\item Specify adaptation/retraining, information timing, and allowed state-update channels before future outcomes.
\item Persist procedure ID and each realized state/artifact/input snapshot.
\item Continue a procedure-level prequential record across allowed updates.
\item Declare a new procedure-level boundary for discretionary rule revisions; preserve earlier evidence under its original object.
\end{enumerate}

\subsection{Incident protocol and compatible alternatives}
\begin{enumerate}[leftmargin=*]
\item Record the direct outcome without automatically labeling it as an internal fault.
\item Declare comparator, anchor, horizon, and scope; label retrospective selections.
\item Resolve which procedure/state generated recommendations and executed actions; list unresolved archive-compatible alternatives.
\item Use telemetry/comparison for subsystem claims and intervention or justified counterfactual design for causal claims.
\item State which conclusions are not established and the evidence needed to distinguish alternatives. If outcomes improve, retain unresolved hypotheses and record the evidence supporting incident closure. Monitoring and improvement do not themselves establish a cause or repair.
\end{enumerate}
These are prospective proposals, not evaluated interventions or historical compliance. They make future evidence more interpretable without claiming to solve causal diagnosis.

\subsection{RQ3 answer}
\begin{quote}
\textbf{RQ3 finding.} Frozen-instance, adaptive-procedure, and execution-state claims need distinct identity bindings. Pre-specified adaptation permits a continuing prospective record; ad-hoc rule changes create new boundaries. Resolvable recommendation-to-runtime and action-to-recommendation links are a practical first priority, not an automatic increase in claim level. Provenance constrains compatible referents, while reference design and causal identification determine how far outcomes support diagnosis.
\end{quote}

\section{Discussion}\label{sec:discussion}

\subsection{A strong oracle can still be a weak debugger}

An external outcome constrains what can be claimed about the realized account path. It does not supply a test assertion naming a violated internal invariant. Financial outcomes are stochastic, reference-dependent, and path-dependent. This separates the value of reality-grounded feedback from its causal resolution.

Software testing already recognizes the oracle problem and the separation between terminal correctness information and fault localization \cite{barr2015oracle,park2025flex}. The present case adds an evolving-system complication: even if the external outcome is accepted as informative, the identity of the evaluated program may itself be moving.

In the present analysis, the small account's historical market-price path is treated as approximately given. This modeling choice does not eliminate order-dependent execution effects or the action-dependent evolution of inventory and cash. In other settings, decisions can also change subsequent data distributions, rewards, or state transitions, as studied in performative prediction and reinforcement learning \cite{perdomo2020performative,mandal2023performative}. Those feedback paths add issues for an identification design to address; this case neither measures their size nor establishes a quantitative amplification law.

\subsection{System-level diagnostic credit assignment}

Temporal credit assignment studies how delayed consequences should be associated with previous actions \cite{pignatelli2024credit}. In a production software/ML stack, the targets of attribution can include models, fitted artifacts, training windows, data paths, ensemble rules, execution decisions, and human overrides. We use \emph{system-level diagnostic credit assignment} as a descriptive phrase for this enlarged attribution problem, not as a replacement for established RL terminology.

\subsection{Technical debt and mutable ML systems}

Sculley et al. describe production ML as vulnerable to entanglement, hidden feedback loops, data dependencies, configuration issues, and changes in the external world \cite{sculley2015debt}. The \sys{} incident is consistent with that systems view: the behavior of a deployed pipeline cannot be reduced to a model coefficient vector. A performance record detached from the identity and lineage of the surrounding stack may answer an organization-level question while failing to answer a model-level one.

\subsection{Provenance and the referent of a claim}

W3C PROV provides a general vocabulary for entities, activities, agents, and derivations \cite{w3c2013prov}; PROV-AGENT extends provenance toward agent-centric prompts, responses, and decisions \cite{souza2025provagent}. A resolved provenance relation can establish the referent of an outcome claim: the procedure, artifact, data state, and action being analyzed. Causal interpretation then requires comparison and identification structure beyond that binding.

Without that binding, causal analysis may be asking the wrong question about the wrong system.

\subsection{Controlled fault injection versus uncontrolled reality}

AIOpsLab represents a complementary regime: controlled environments, injected faults, generated workloads, and telemetry for evaluating incident-management agents \cite{chen2025aiopslab}. Control over fault generation supplies known experimental structure for diagnosis.

Naturally occurring cloud incidents offer a further comparison: RCACopilot aggregates runtime information and produces root-cause categories and explanatory narratives \cite{chen2024rcacopilot}. Its incident records and operational labels provide diagnostic evidence that our market case does not inherit; a category prediction is not by itself an identified causal mechanism. The present case trades controlled clarity for ecological validity. The market cannot be reset; failures are naturally occurring; action consequences are entangled with the environment; and the system was being modified during deployment. The lesson is not that live evidence is superior to controlled benchmarks, but that causal certainty from controlled benchmarks should not be projected onto uncontrolled longitudinal production data.

\subsection{The rebound is also epistemically constrained}

The same evidentiary discipline applies to favorable and unfavorable outcomes. The selected July--September segment returns +15.2461\% over 65 observations; the single-market annualized intercept is +76.5417\% ($p=0.00230$), and the multi-proxy intercept is +53.3570\% ($p=0.00793$). These are retrospective, specification-dependent estimates. July and August improved, while September was negative; candidate results also vary by horizon and class. Incomplete runtime identity prevents the archive from establishing that the same frozen model recovered or that a particular intervention repaired it.

Improvement can reduce the perceived urgency of investigation, but it should not substitute for resolving the original diagnostic questions. Work on automation-related complacency and bias provides a broader reason to attend to oversight of imperfect automated aids \cite{parasuraman2010complacency}; we did not measure those psychological effects or show that this operator stopped investigating. The prospective recommendation is to record what identity and comparative evidence supports closure, and which hypotheses remain open. Good outcomes can leave the same identity and causal questions unresolved.

\subsection{Why the approximate random reference is useful anyway}

The June approximate random reference is still useful because it answers a limited descriptive question: where did the observed production result sit relative to uninformed selection under an approximate common portfolio rule? It does not need to be a strict hypothesis test to provide context.

The annual approximate reference is equally important. Reporting both the poor selected month and the strong annual position helps prevent the benchmark itself from becoming a cherry-picking device. A broader methodology for multiple null and perturbation families and benchmark design belongs to a separate study rather than being consumed here.

\subsection{Execution friction as bounded context}
Execution was measurably imperfect: annual notional-weighted price friction differs by side, and fills include manual and out-of-pool actions. Appendix~\ref{app:friction} retains definitions and values. This annual measurement is not an incident decomposition and cannot exclude execution as an explanation. Price friction is not deducted again from account returns that already include actual settlements.

\subsection*{\textit{Meanwhile, in the passive pool\ldots}}
Over the common history from 1 July 2024 to 30 September 2026, a separate allocation to cash, an equity-index fund, bonds, and gold recorded an approximate cashflow-neutral cumulative return of 30.64\%, compared with 27.75\% for the quantitative account.\footnote{Both cumulative returns use the 28 June 2024 closing baseline. This inception-to-cutoff context extends beyond the main Paper-II study year. Passive returns approximately chain recorded NAV changes after external-flow adjustment; the separate pool is outside S1.} The pool was maintained through contributions and approximate rebalancing, and also enjoyed a strong stretch while the quantitative system was being revised and debugged. Its longer record was more modest: the approximate cashflow-neutral calendar CAGR from 1 October 2021 to 30 September 2026 was 5.93\%. The passive pool is shown only as informal historical context; it is not a matched, risk-adjusted, or causally interpretable benchmark.

\section{Related Work}\label{sec:related}

\subsection{Coding agents and human oversight}
SWE-bench evaluates issue-driven repository changes, and SWE-agent supplies interfaces for repository navigation, editing, and program execution \cite{jimenez2024swebench,yang2024sweagent}. Our longitudinal question adds runtime identity, human acceptance, and action binding to this task-level view of engineering. In this case, tasks, direction, and deployment oversight remained human-controlled. Human--automation research discusses complacency and bias in the use of imperfect aids \cite{parasuraman2010complacency}; it motivates an oversight question, not a diagnosis of the operator's psychology.

\subsection{Oracles, fault localization, and diagnostic justification}
Test-oracle research compares behavior with an acceptance specification \cite{barr2015oracle}. Delta debugging isolates failure-inducing inputs through repeated tests \cite{zeller2002delta}; external-oracle localization uses counterfactual execution \cite{park2025flex}. These separate terminal feedback from causes. Our focus is the identity of the evaluated object when a deployed procedure evolves and historical stochastic conditions cannot literally be rerun.

JustDiag tracks evidence, findings, competing hypotheses, conflicts, and next checks, separating answer quality from process quality \cite{bi2026justdiag}. The present case complements this process view by asking how action-to-runtime identity bounds acceptance of a longitudinal diagnostic claim.

\subsection{Credit assignment and adaptive evaluation}
Temporal credit assignment associates delayed consequences with prior actions, including noisy or sparse feedback \cite{pignatelli2024credit}. Our diagnostic targets also include code, artifacts, ensemble policies, and human execution. The target here is identity-bound system attribution in longitudinal production.

Dawid's prequential view permits sequential assessment through observations arriving after prediction \cite{dawid1984prequential}. Concept-drift work studies adaptation and evaluation when predictive relationships change \cite{gama2014drift}. Neither parameter variation nor adaptation alone invalidates prospective evaluation. We distinguish state evolution under a declared procedure from discretionary procedure revision and show why that distinction needs runtime binding.

Performative prediction studies decision-induced changes in the data distribution \cite{perdomo2020performative}; performative RL considers policy-dependent rewards and transition dynamics \cite{mandal2023performative}. They provide related frameworks for endogenous feedback, not a theorem about the diagnostic difficulty of this account. Our empirical comparisons treat the observed market path as approximately given and do not estimate such feedback effects.

\subsection{Production ML observability, technical debt, and lineage}
Observability addresses detection, diagnosis, and reaction across pipelines with silent failures \cite{shankar2021observability}. Technical-debt work identifies entanglement, configuration issues, hidden feedback loops, and changing dependencies \cite{sculley2015debt}. Data cascades describe downstream effects of upstream practices that may be delayed or difficult to observe \cite{sambasivan2021cascades}. These motivate pipeline diagnosis. We ask which versioned scientific object a changed production metric evaluates.

ModelDB tracks and queries models/pipelines across iterative development \cite{vartak2016modeldb}. PROV-DM represents general provenance and PROV-AGENT traces agent interactions \cite{w3c2013prov,souza2025provagent}. TraceCaps proposes step-level provenance capsules and inline risk gating for agentic software engineering, with early demonstrations \cite{catarino2026tracecaps}. Its focus is engineering-step control; our proposed binding links recommendations to runtime identities and subsequent realized actions. These registries and lineage standards provide context for the proposed manifest, which connects procedure/artifact lineage to information timing, substitutions, and realized actions so that outcome claims have an explicit referent.

\subsection{Controlled incidents and causal identification}
AIOpsLab integrates cloud environments, injected faults, workloads, telemetry, and interfaces for evaluating incident-management agents \cite{chen2025aiopslab}. Known fault-generation structure helps evaluation. RCACopilot instead studies naturally occurring cloud incidents using runtime information and operational root-cause categories \cite{chen2024rcacopilot}. Our case lacks both an experimenter-supplied root-cause label and an equivalent labeled incident corpus, as well as a resettable market path. This is a complementary setting, not evidence that live observation is superior or all benchmarks are perfectly controlled.

Statistical fault localization also faces confounding: Baah et al. use causal modeling to distinguish execution associations from effects under explicit assumptions \cite{baah2010causal}. Causal inference requires identification assumptions beyond association \cite{hernanrobins2020}. Knowing what ran resolves identity, not its causal effect. The compatible-history criterion organizes this distinction rather than providing a new causal estimator.

\subsection{Financial selection and inference}
Deflated Sharpe Ratio and Probability of Backtest Overfitting address selection-driven inference \cite{bailey2014dsr,bailey2017pbo}; multiple-testing work warns against treating searched financial findings as isolated tests \cite{harvey2016multiple}. HAC addresses residual covariance, not window selection \cite{neweywest1987}. We do not compute DSR, PBO, or a selection-adjusted June test. Selected-period caveats, specification sensitivity, explicit approximate references, and annual context remain essential to the case.

Gelman and Stern explain why comparing significance labels does not test the difference between estimates \cite{gelman2006significance}. We apply that caution to the market-only and multi-proxy intercepts rather than treating their different $p$-values as an identified model contrast or environmental explanation.

The case connects these established methods and warnings to the interaction of outcome evidence, mutable evaluated-object identity, incomplete provenance, and causal claim granularity in an agent-engineered system.

\section{Threats to Validity}

\subsection{Retrospective selection}

The annual window was fixed for this retrospective analysis; the segmented windows and June focus followed observation of the trajectory. HAC covariance does not correct window selection or multiple testing. Selected-window estimates and Monte Carlo lower fractions therefore do not have a preregistered interpretation. The diagnostics draw on a shared archive and are not independent replications.

\subsection{Reference, replay, and accounting approximations}

The subset-EW reference is gross and idealized, with membership, delisting, and carried-price conventions; candidate-relative results use signal-close matching only. Opening replays and random references simplify manual tradability judgments, substitutions, integer lots, cash allocation, and fees, including minimum-fee mechanics and small cash imbalances. The random reference follows a five-trading-day schedule on a continuous annual path before slicing June. Fixed-basket and account-linked contributions answer different accounting questions, and neither is FIFO realized P\&L. These constructions support scoped comparisons rather than exact alternate histories or root-cause probabilities.

\subsection{Archive completeness and deployment binding}

Repository and configuration dates do not establish activation dates or recommendation-to-runtime links. The limited run/cycle inventory is not an annual action census, and success declarations are not independent verification. Omitted actions, artifacts, overrides, or failed logging may constrain resolution further. Missing bindings do not show that particular alternative loaders ran, and no inferred deployment epochs fill those gaps. The illustrative histories in Section~\ref{sec:compatible} explain the criterion; they are not checked witnesses for this incident.

\subsection{Statistical and model specification}

Regression inference depends on the stated covariance, risk-free-rate, and explanatory-variable specifications. HAC(3) is a retrospective presentation convention, with OLS, HAC(5), and HAC(10) sensitivities retained; the style proxies are not canonical pricing factors. Significance levels do not determine which specification is retained. Repeated candidate occurrences and overlapping horizons are not independent samples, and regression HAC does not calibrate their descriptive summaries. Right-censoring at the cutoff requires observed coverage to be reported rather than assumed complete.

\subsection{Single-case generalization and protocol status}

This is one system and one domain. Agent-assisted implementation and human authority are operator-reported and partly illustrated by notes, rather than quantified through authorship, modification speed, or oversight measures. Finance supplies stochastic outcomes, delayed attribution, execution constraints, and a historical path that cannot be reset; other settings may offer controlled replay or decision-induced environmental feedback. The case does not estimate that feedback or establish population-level magnitudes or a complexity-growth law. The proposed production-binding protocol was not historically implemented in full and is not prospectively validated by this case; its diagnostic benefits remain to be evaluated.

\section{Conclusion}

The \sys{} case separates realized outcome evidence, evaluated-object identity, and causal diagnosis. Across the fixed-for-analysis 2025-10-01--2026-09-30 annual window, the account gained and outperformed CSI300, while annual alpha was not statistically distinguishable from zero under the main retrospective specification. Selected subperiods include severe adverse relative performance and strong later outcomes. Candidate and random-reference diagnostics provide conditional descriptions under declared scopes; style, cohort, and replay analyses constrain specific narratives.

Engineering/configuration changes overlap the episode, but complete historical action-to-runtime binding is unavailable. The archive therefore supports an evolving account record rather than establishing one frozen instance or a unique cause. The later gains do not establish recovery of the same model. This \emph{outcome--diagnosis gap} concerns both the referent and the explanation of feedback, complementing Paper I's emphasis on external evaluation constraints.

For longitudinal evaluation, the first task is to make the outcome's referent resolvable. Frozen-instance claims require material identity binding; a pre-specified adaptive procedure can maintain its prospective record across declared state updates. Separate links from recommendations to runtime artifacts and from recommendations to executed actions preserve these distinctions.

The resulting principle is:
\begin{quote}
\textbf{Reality may identify that an outcome occurred without identifying which evolving procedure produced it or why.}
\end{quote}

The proposed binding protocol specifies evidence for that factual task and remains prospectively unvalidated. Reference design, comparative evidence, and causal assumptions determine how far resolved identity can support diagnosis.

\section*{Acknowledgments, Funding, and Declarations}

\noindent\textbf{Author Contributions:}
\textbf{Kun Liu} led the study design, retrospective analysis,
development of the evaluated-object identity and evidence framework,
and preparation of the manuscript. \textbf{Liqun Chen} led live
market execution and contributed financial-domain expertise and
operational interpretation.

\medskip
\noindent\textbf{AI Assistance and Human Responsibility:}
Large language models and coding agents assisted with software
engineering, diagnostic analysis, experiment preparation, and
manuscript drafting and revision under human-issued tasks. They are
not listed as authors and had no independent authority to initiate
production changes, deploy software, or execute trades. The human
authors reviewed the evidence, analyses, and manuscript, retained
deployment and trading authority as described in the paper, and take
responsibility for the reported claims.

\medskip
\noindent\textbf{Funding and Resource Allocation:}
This research was self-funded. Computational resources and research
expenses were provided privately by the authors. The live trading
account used private capital contributed by both authors, with most
initial trading capital supplied by \textbf{Liqun Chen}.

\medskip
\noindent\textbf{Competing Interests:}
The authors developed and operated the evaluated system and had a
direct financial interest in the studied trading account, benefiting
from gains and bearing losses.

\medskip
\noindent\textbf{Acknowledgments:}
We acknowledge the anonymous market participants whose uncompromising
``peer review'' supplied the live external feedback examined in this
study. We dedicate this work to Xing Liu, our most cherished
``long-term alpha.''

\medskip
\noindent\textbf{Code and Data Availability:}
The QuantPits engine and project documentation are available at
\url{https://QuantPits.com}. The public engine is not a complete
replication package for the empirical analyses reported here.
Supplementary Dataset S1 (v1) is available on Zenodo at
\url{https://doi.org/10.5281/zenodo.23215521} under the Creative Commons
Attribution-ShareAlike 4.0 International license.
Appendix~\ref{app:availability} describes its contents, provenance, and
limitations. Raw brokerage records and private runtime evidence remain withheld.

\medskip
\noindent\textbf{Disclaimer:}
This study is presented for research purposes and does not constitute
investment advice, a recommendation, or a solicitation to trade any
security. Deployment of agent-engineered trading software can incur
substantial financial losses. The reported historical results do not
establish persistent investment alpha or future profitability.

\clearpage
\appendix
\section{Complete Monthly Numerical Context}\label{app:monthly}
Table~\ref{tab:monthly} and Figure~\ref{fig:monthly_returns} report the same contemporaneous compounded monthly returns.

\begin{table}[htbp]
\centering
\caption{Contemporaneous compounded monthly returns. EW is the approximate gross subset reference; $n$ counts daily account returns.}
\label{tab:monthly}
\small
\begin{tabular}{lrrrr}
\toprule
Month & $n$ & Account & CSI300 & Approx. EW \\
\midrule
2025-10 & 17 & +3.0224\% & -0.0004\% & +1.5938\% \\
2025-11 & 20 & -1.6414\% & -2.4568\% & -2.3050\% \\
2025-12 & 23 & +4.3952\% & +2.2816\% & +2.0311\% \\
2026-01 & 20 & +5.4351\% & +1.6501\% & +2.7022\% \\
2026-02 & 14 & +1.4690\% & +0.0916\% & +2.2296\% \\
2026-03 & 22 & -4.2922\% & -5.5321\% & -6.0845\% \\
2026-04 & 21 & +1.8440\% & +8.0282\% & +3.9476\% \\
2026-05 & 18 & -3.6901\% & +1.7642\% & -3.4612\% \\
2026-06 & 21 & -9.2297\% & +1.7847\% & -4.8508\% \\
2026-07 & 23 & +9.8920\% & -7.8569\% & +2.7974\% \\
2026-08 & 21 & +6.1394\% & +0.8040\% & +0.8910\% \\
2026-09 & 21 & -1.1940\% & -5.7830\% & -4.0809\% \\
\bottomrule
\end{tabular}
\end{table}

\section{Secondary Candidate, Attribution, and Replay Diagnostics}\label{app:secondary}
These measurements support the diagnostic questions in Section~\ref{sec:diagnostic-narratives}. Horizon, recurrence, Sell, and Swap summaries constrain claims about universal failure, freshness, and replacement. Cohort accounting distinguishes the fixed basket from the actual account; opening replay and freeze valuations test specific retained-holdings narratives under stated approximations. Each answers a bounded question rather than supplying a root-cause label. Definitions and numerical evidence follow the methods and cutoff in Section~\ref{sec:system}.

\subsection{Candidate outcomes across horizons}

The signal-close and execution-open curves separate primary and all candidates (Appendix~\ref{app:diagnostic-figures}). Outcomes are mixed across horizons: April's short horizons are not uniformly positive, June $h1$/$h3$ are positive in some scope/anchor combinations, and July does not show universal improvement. The plots summarize absolute candidate returns rather than realized trade P\&L. Per-cell requested/observed counts and cutoff statuses are retained in the local research archive; they are not assumed identical across horizons. These forward-return curves do not estimate an information half-life.

\subsection{Candidate persistence}

The annual archive contains 483 candidate-occurrence records for the persistence analysis. For 422 occurrences (87.3706\%), the running consecutive-batch count is one: the stock was absent from the preceding recorded batch, or the occurrence is in the first recorded batch. This statistic counts starts of runs, including runs that later continue; it is not the fraction of complete sequences whose final length is one. The longest observed consecutive run is five recorded recommendation batches, and the archive contains 195 unique Buy-candidate stocks.

This supports a narrow statement about recommendation recurrence: most occurrences begin a run rather than continue from the immediately preceding recorded batch. It does \emph{not} prove that the underlying model artifacts, data, features, or latent signals were fresh, and it does not by itself eliminate stale-data or stale-artifact explanations.

\subsection{Sell-side observations}

Suggested Sell candidates are measured from intended execution-day open to horizon close, not from actual realized sale P\&L. At T+10, the April median ($n=12$) is +0.7232\% with 66.67\% positive, whereas the June median ($n=15$) is $-4.2904$\% with 40\% positive. These values do not identify Sell logic as the primary cause, but neither do they justify claiming that Sell-side effects have been excluded.

\subsection{Swap observations}

The swap diagnostic forms unique-stock baskets from same-day actual Buy and Sell fills. Within each side it takes the equal-weight mean stock return from execution-day open to the shared horizon close, then subtracts Sell from Buy. Monthly summaries are medians of these batch differences; the positive fraction counts batches with positive differences. It is not a one-to-one matched-trade design, not realized P\&L, and does not isolate opportunity cost or manual execution effects.

At T+10, March, April, May, and June have 5, 4, 4, and 5 fully observed batches, respectively:
\begin{center}
\begin{tabular}{lrr}
\toprule
Month & Median Buy-minus-Sell (pp) & Positive fraction \\
\midrule
March & +1.3288 & 60\% \\
April & -3.1172 & 0\% \\
May & +0.7242 & 50\% \\
June & -1.0081 & 40\% \\
\bottomrule
\end{tabular}
\end{center}
The mixed pattern does not establish a uniformly negative replacement mechanism.

\subsection{Fixed-basket versus account-linked contribution}

We distinguish a fixed equity basket from actual account-linked attribution.

Cohorts are fixed at the 29 May holding snapshot. Labels reflect the start of the stock's uninterrupted positive-holding episode in snapshots available by that date, not FIFO acquisition lots; episodes already present at the beginning of the available archive are left-censored. A \emph{fixed-basket contribution} uses 29 May equity-value weights, excluding cash, and holds adjusted units fixed through 30 June. It is not the actual June NAV decomposition. Under that construction, the May-entry basket has a median June return of $-8.0756$\%, a mean of $-10.0925$\%, and a fixed-basket contribution of $-4.8413$ pp.

The \emph{account-linked contribution} follows daily stock valuation changes plus signed settlement cash, including recorded fees and dividend tax; these costs are already in settlements and are not subtracted again. Absent stock rows represent zero holdings only within verified complete daily snapshots. Daily stock P\&L divided by cash-flow-adjusted account capital is multiplied by preceding cumulative June wealth and summed, so contributions reconcile to actual June account return.

The May-labeled group contributes $-2.84050$ pp, the April-labeled group $-1.40736$ pp, and stocks outside the fixed May-29 cohort $-5.06181$ pp. March-or-earlier stocks contribute +0.07994 pp, cash-only settlements approximately +0.000043 pp, and the reconciliation residual is numerically negligible. The sum of unrounded components matches June's $-9.2297$\% account return, displayed to four decimal places. The outside-cohort group is the largest negative contribution; it is neither the whole remainder of the May-29 basket nor necessarily all June acquisitions. May-labeled stocks were not the largest negative account-linked contribution, and cohort membership does not establish causal responsibility.

\begin{figure}[htbp]
\centering
\includegraphics[width=0.96\linewidth]{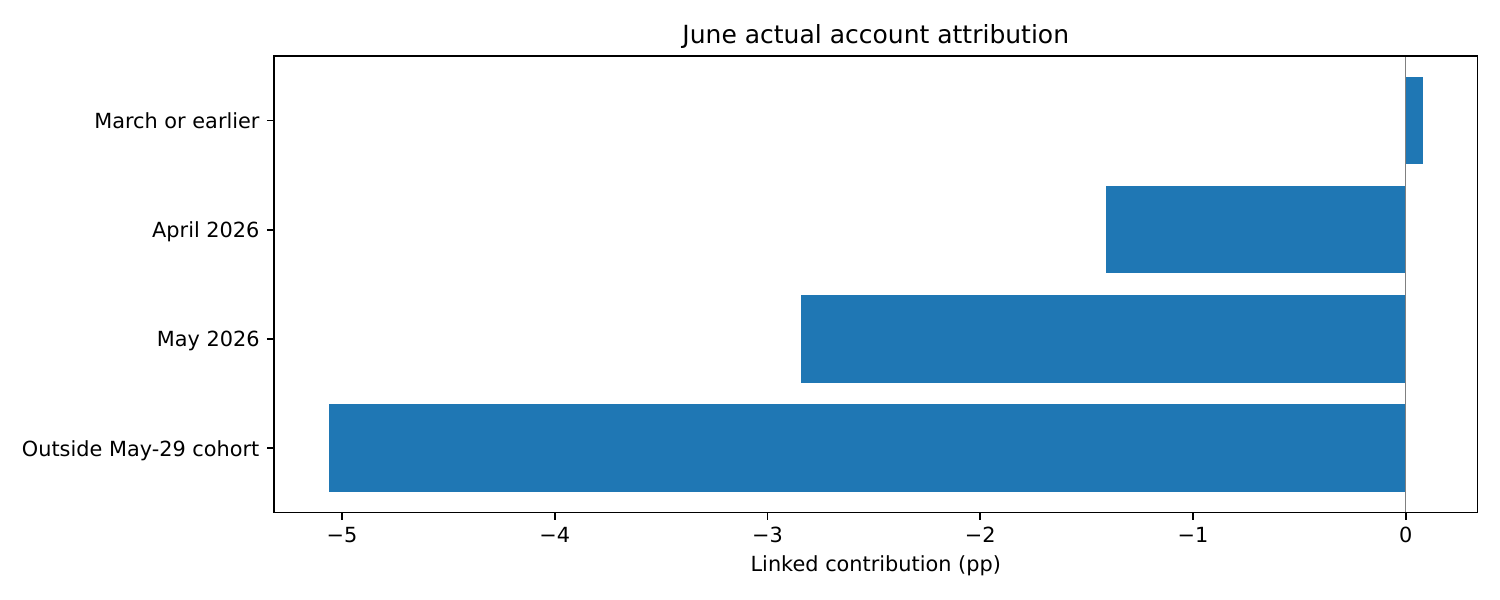}
\caption{Wealth-linked June account return contributions, using fixed May-29 instrument cohorts and signed settlements. Cohorts are ordered by holding-episode date, followed by stocks outside the May-29 cohort; the latter are not all June acquisitions. Cash-only settlements (+0.000043 pp) and the negligible reconciliation residual are retained in the full decomposition but omitted from this plot. This is not FIFO realized P\&L.}
\label{fig:june_account_contributions}
\end{figure}

\subsection{Opening-policy replay}

The opening replay is an approximation, not a reconstruction of actual manual execution. It uses integer 100-share buys, sell-first sequencing, approximate cash equalization, candidate substitutions, retained cash, and the configured fee approximation (buy rate 0.0005; sell rate 0.0015; minimum fee 5). Actual historical opening tradability involved human judgment, and ex-post opening price changes are not exchange-state labels.

The replay initializes the 22 actual holdings and cash at the 2 March close and trades only on subsequent recorded recommendation dates through 30 June. Position counts can decline if replacement buys cannot be filled. The opening-gap proxy blocks buys in either direction at an absolute adjusted open/previous-close gap of at least 9.5\%, and sells at a downward gap of at least 9.5\%; it is not historical exchange limit metadata. Recommended held-stock exits are prioritized; when a larger DropN needs additional exits, a deterministic instrument-order heuristic selects held stocks outside the Buy list. This does not reconstruct a complete historical score ranking. Buy budgets are approximately equal across available slots; fees are reserved before rounding down to whole lots, and unfilled slots retain cash. Initial adjusted units remain fixed between trades, cash earns no interest, and no external funding is added.

Approximate replay returns are $-16.6910$\% for Drop0 and $-19.6462$\% for Drop1. Drop3, Drop6, and Drop9 return $-12.9816$\%, $-11.8967$\%, and $-17.0115$\%, respectively. Drop6 exceeds configured Drop3 on this approximate path, but the non-monotonic comparison does not establish an optimal live policy or show that changing turnover would have fixed the incident. Appendix~\ref{app:diagnostic-figures} shows the paths and summary.

\subsection{Freeze counterfactual}

Freeze valuations hold initial adjusted units and initial cash through 30 June, with cash earning zero and no subsequent trades. The three requests are retrospectively specified snapshot sensitivities, not a uniform calendar-month-start design. According to the operator, the 2 March closing snapshot was chosen for the follow-on analysis after Paper I's 27 February endpoint; it also initializes the opening-policy replay. This research choice was unrelated to the partial discretionary cash-out. The baseline for each specified request is the latest available closing snapshot on or before that date.

Within each row, both columns use the same interval strictly after the actual valuation baseline and through the 30 June close. The 2 March-close comparator therefore contains 81 daily return observations; the full 82-day March--June segment also includes 2 March's return from the 27 February close. The rows have different holdings and durations. Table~\ref{tab:freeze} reports the actual baselines.
\begin{table}[htbp]
\centering
\caption{Approximate fixed-unit freeze counterfactuals from the stated baseline closes to the 30 June 2026 close. Initial cash is included. Both columns exclude the baseline date's return; $n$ counts subsequent daily production-return observations, not holdings.}
\label{tab:freeze}
\small
\begin{tabular}{lrrrr}
\toprule
Baseline close & $n$ & Frozen basket & Production & Frozen minus production \\
\midrule
2 Mar 2026 & 81 & -16.6910\% & -15.4779\% & -1.2131 pp \\
1 Apr 2026 & 59 & -7.2509\% & -12.0480\% & +4.7970 pp \\
30 Apr 2026 & 39 & -11.9045\% & -12.5792\% & +0.6747 pp \\
\bottomrule
\end{tabular}
\par\smallskip
\begin{minipage}{0.96\linewidth}
\footnotesize
First included return dates are 3 March, 2 April, and 6 May, respectively. Requests were specified retrospectively as 2 March, 1 April, and 1 May, rather than by a uniform calendar-month-start rule. Each request uses the last available close on or before its date; the 1 May request therefore uses 30 April. Both columns exclude the baseline date's return.
\end{minipage}
\end{table}

The 1 April snapshot's freeze exceeds production over its matched interval. This is compatible with harmful subsequent production decisions, but does not prove a causal failure onset or identify the responsible component. The three rows have different holdings, start dates, and durations; their cumulative returns do not identify an optimal freeze date.

\section{Approximate Random-Reference Mechanics}\label{app:random-methods}
We use the engine's pure numerical TopK Dropout random-ranking function with $K=22$ and Drop0/3/22, generating 5,000 continuous annual paths per profile with seed 20261001. Profiles share each trial's uniform random scores. Every five trading days, scores rank the prior-day eligible subset; Drop3 removes low-ranked holdings and replenishes from eligible candidates, with forced exits for out-of-pool holdings on rebalance dates. June is sliced from the annual paths rather than restarted at actual June holdings. Daily returns are ideal equal-weight means even between scheduled constituent changes, using historical carried-price valuation as in the approximate EW reference. Suspended stocks can consequently enter simulations despite being non-executable. After initialization, the new-buy count divided by $K$ is multiplied by the combined configured buy/sell rate (0.002) and deducted on rebalance dates. No initial fee, minimum fee, whole-lot, opening-fill, or manual-override mechanics are modeled. These simplifications define an approximate random reference rather than a matched production null.

\section{Factor-Proxy Construction and Sensitivity}\label{app:factor-methods}
Multi-factor sensitivity adds three lagged cross-sectional long--short proxies to market excess return: log adjusted-price times volume, 20-trading-day momentum, and 20-trading-day return volatility (minimum five observations). Each characteristic is lagged one trading day, with a 45-calendar-day buffer before the first analyzed day. On each day's prior-day subset members, the common complete observed cross-section is used without filling stock prices; each factor is the mean current-day return at or above its characteristic's 80th percentile minus the mean at or below its 20th percentile. At least five complete stocks are required. These are engine-inspired liquidity, momentum, and volatility proxies, not canonical Fama--French, Barra, or industry-neutral factors. Annual, initial, incident-background, and later regressions use 241, 94, 82, and 65 days, respectively, under the same covariance/risk-free conventions. The reconstructed definition differs from the legacy production factor implementation; this sensitivity is documented in the local research archive.

\begin{table}[htbp]
\centering
\caption{Market-only covariance sensitivity, with annual $r_f=1.35\%$. Windows match Table~\ref{tab:periods}; no model is refitted for this display.}
\label{tab:covariance-sensitivity}
\small
\begin{tabular}{lrrrrrr}
\toprule
Window & $n$ & Annual $\alpha$ & OLS $p$ & HAC(3) $p$ & HAC(5) $p$ & HAC(10) $p$ \\
\midrule
Full year & 241 & +14.4911\% & 0.28244 & 0.27526 & 0.30006 & 0.32239 \\
Initial & 94 & +30.0744\% & 0.09396 & 0.03339 & 0.02222 & 0.00893 \\
Mar--Jun & 82 & -58.1058\% & 0.02268 & 0.01778 & 0.02319 & 0.02642 \\
Jul--Sep & 65 & +76.5417\% & 0.00242 & 0.00230 & 0.00349 & 0.00473 \\
\bottomrule
\end{tabular}
\par\smallskip
\begin{minipage}{0.96\linewidth}
\footnotesize
Within each row, returns, regressor, risk-free convention, sample, and intercept are identical; covariance estimation changes uncertainty. HAC uses Bartlett weights, small-sample correction, and a $t$ reference. Lag 3 is the fixed main presentation convention for this retrospective analysis, with no documented optimal or prospective selection. The Initial segment overlaps Paper I's Phase 3; its OLS values agree at reported precision, while its 5\% significance classification changes under HAC. The other displayed windows retain their 5\% classifications across these settings. These comparisons do not correct retrospective window selection.
\end{minipage}
\end{table}
\FloatBarrier

\section{Execution-Friction Definitions and Full-Year Context}\label{app:friction}

Across the full annual archive, the transaction aggregation reports transaction-notional-weighted Buy total friction of approximately +0.0348\% and Sell total friction of approximately $-0.3839$\%. Excluding the three M-classified Sell fills yields approximately $-0.3810$\% on Sell friction. All 314 fills are retained, including out-of-current-pool stocks: 162 Buys and 152 Sells; excluding M leaves 149 Sells. The aggregate explicit-fee ratio, computed as total recorded fees divided by total transaction notional, is approximately 0.0528\%. M classification is separate from owner annotations and unrecorded feasibility judgments.

The operator reports omitted universe-exit handling, and the records show one delayed exit and one purchase after official index removal before both positions were sold in August; dividend- and tax-aware cash-only comparisons have mixed signs, without identifying these departures' contribution to annual account performance.

For $s=+1$ on Buys and $s=-1$ on Sells, signal-to-open friction is $s(P_{\mathrm{prev}}^{\mathrm{adj}}-P_{\mathrm{open}}^{\mathrm{adj}})/P_{\mathrm{prev}}^{\mathrm{adj}}$ and open-to-execution friction is $s(P_{\mathrm{open}}^{\mathrm{raw}}-P_{\mathrm{exec}})/P_{\mathrm{open}}^{\mathrm{raw}}$. Negative values indicate adverse movement on both sides. Total friction is the arithmetic sum of components with different denominators, not an exact compounded full-interval price change. Price diagnostics exclude the separately reported explicit fee ratio and are not additional cash deductions from NAV.

These are observed execution-friction measurements. This paper does not infer from their scale that execution ``cannot explain'' the broader performance episode; such a causal exclusion would require a formally specified decomposition. The reported friction scope is the full annual archive, not an incident-window decomposition.

\section{Additional Diagnostic Figures}\label{app:diagnostic-figures}

These figures use the same evidence archive as the main tables. Horizon curves report absolute candidate-occurrence medians, with primary and all scopes separated; the local research archive retains observed/requested counts for each cell. The coverage figure shows observed and censored primary signal-close labels. Missing-price statuses are kept separately in the data; in this annual extraction the 230 unobserved candidate labels are all cutoff-censored. Replay and factor results retain the Methods approximations.

\begin{figure}[htbp]
\centering
\includegraphics[width=0.96\linewidth]{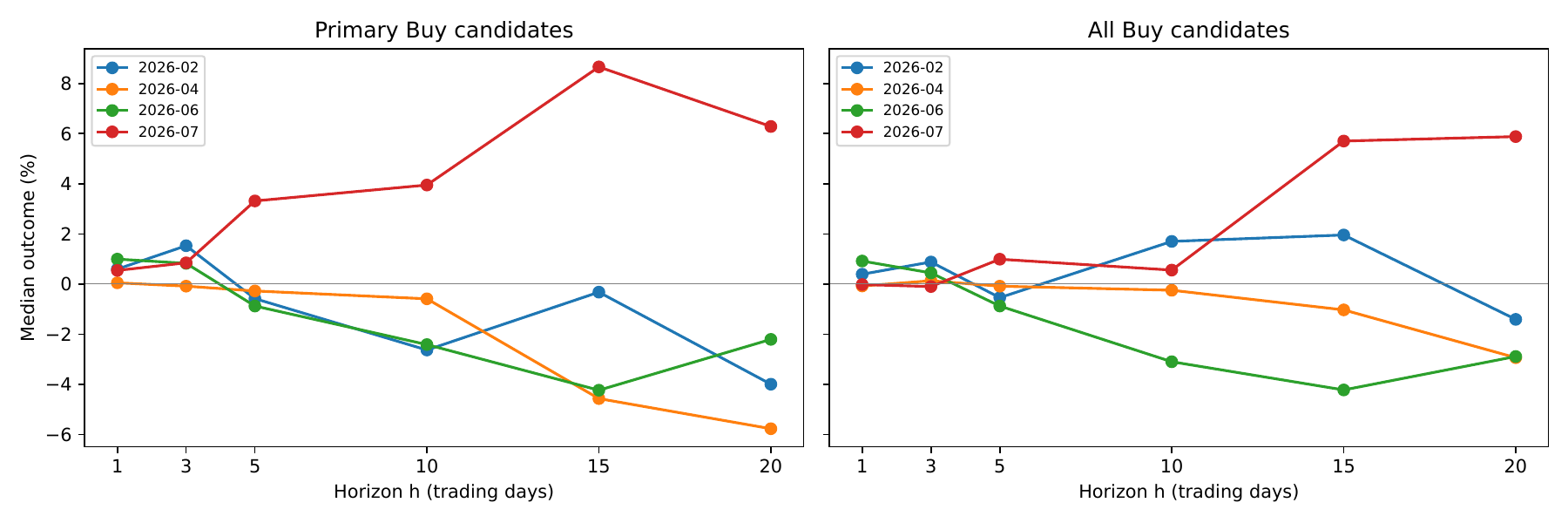}
\caption{Median Buy-candidate returns by horizon from previous trading-day close to target close, for February, April, June, and July; primary and all scopes are separate. Only the six labeled integer horizons are evaluated; $h=1$ ends at the intended execution-day close.}
\label{fig:horizon_signal_close}
\end{figure}

\begin{figure}[htbp]
\centering
\includegraphics[width=0.96\linewidth]{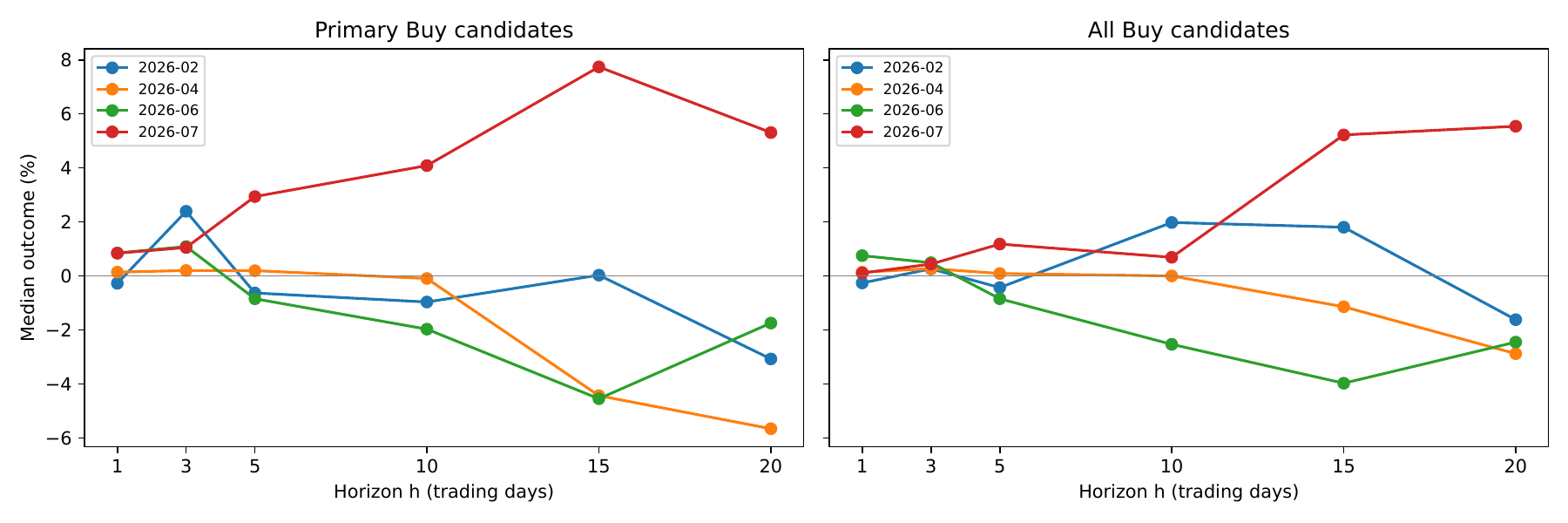}
\caption{Median Buy-candidate returns for the same four months, measured from intended execution-day open to target close. Only the six labeled integer horizons are evaluated; $h=1$ ends at that day's close. No matched execution-open EW reference is used.}
\label{fig:horizon_execution_open}
\end{figure}

\begin{figure}[htbp]
\centering
\includegraphics[width=0.96\linewidth]{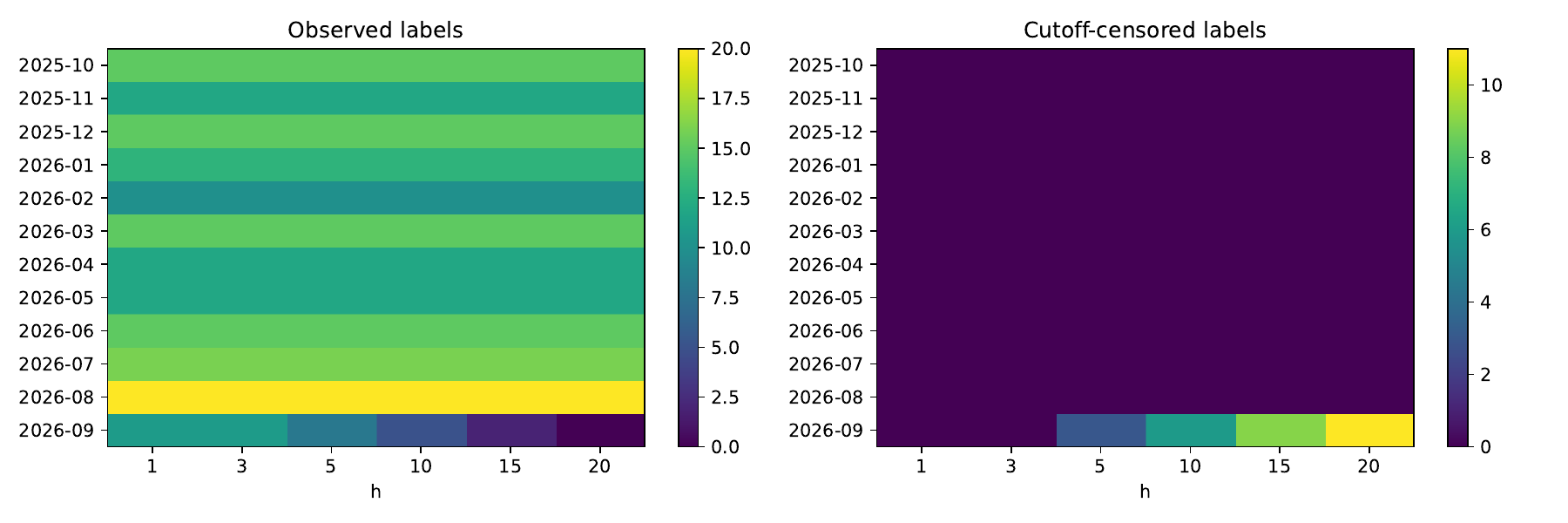}
\caption{Observed and fixed-cutoff-censored primary Buy signal-close labels by month and horizon. Missing prices are distinct statuses in the data, not zero returns.}
\label{fig:candidate_coverage}
\end{figure}

\begin{figure}[htbp]
\centering
\includegraphics[width=0.96\linewidth]{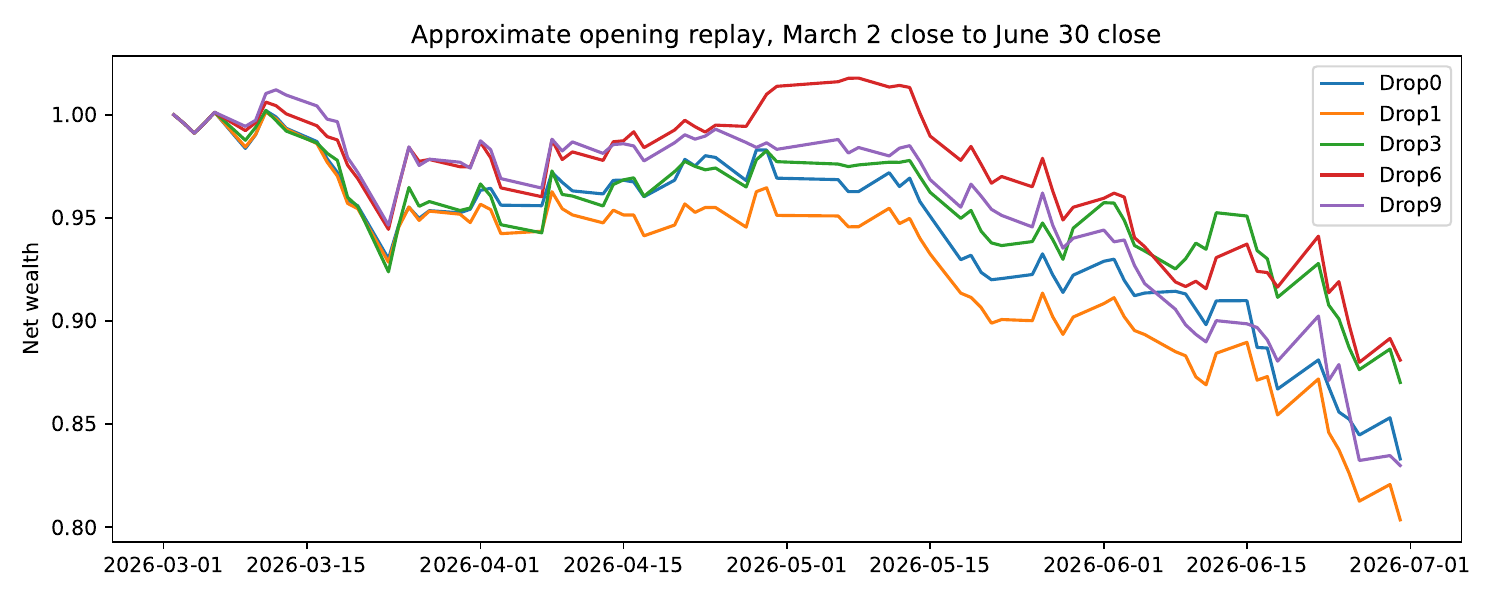}
\caption{March 2-close to June 30 approximate opening-policy wealth paths. Whole-lot buys, opening-gap blocks, heuristic additional exits, retained cash, and configured fees are simulated.}
\label{fig:opening_replay}
\end{figure}

\begin{figure}[htbp]
\centering
\includegraphics[width=0.96\linewidth]{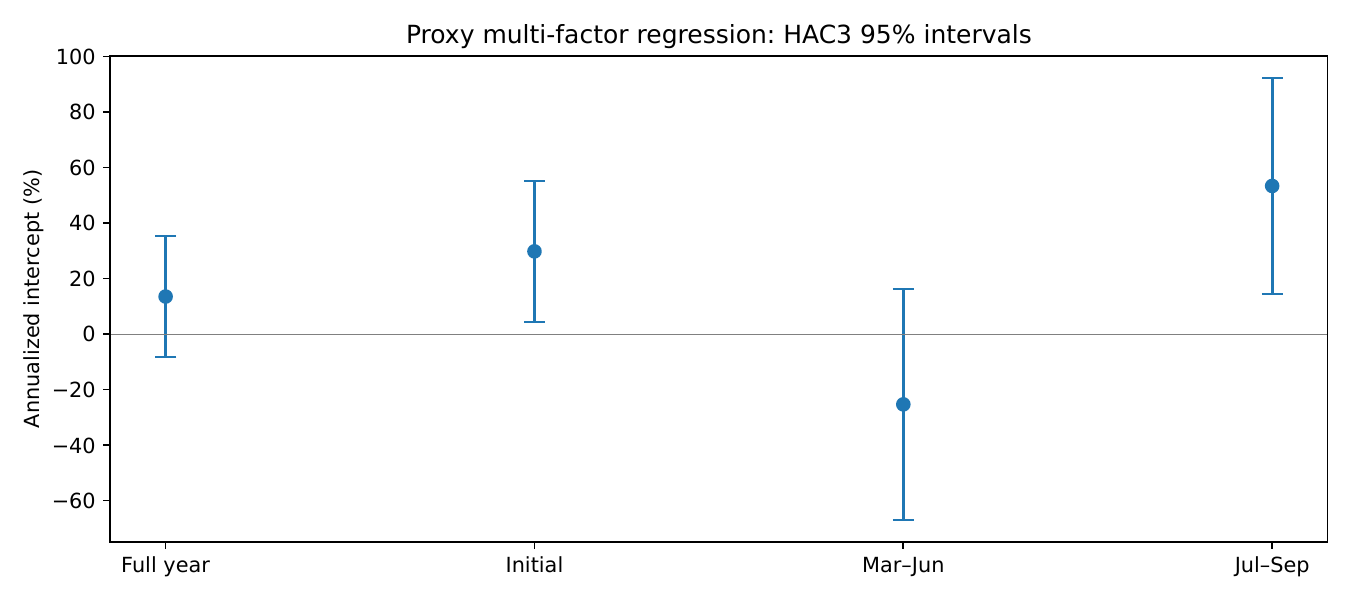}
\caption{Annualized multi-factor proxy intercepts with 95\% HAC(3) confidence intervals. Labels match Table~\ref{tab:periods}; segments are retrospective. Proxies use prior-day members and unfilled observed stock prices. The Mar--Jun interval spans zero under this specification.}
\label{fig:factor_intervals}
\end{figure}

\begin{figure}[htbp]
\centering
\includegraphics[width=0.96\linewidth]{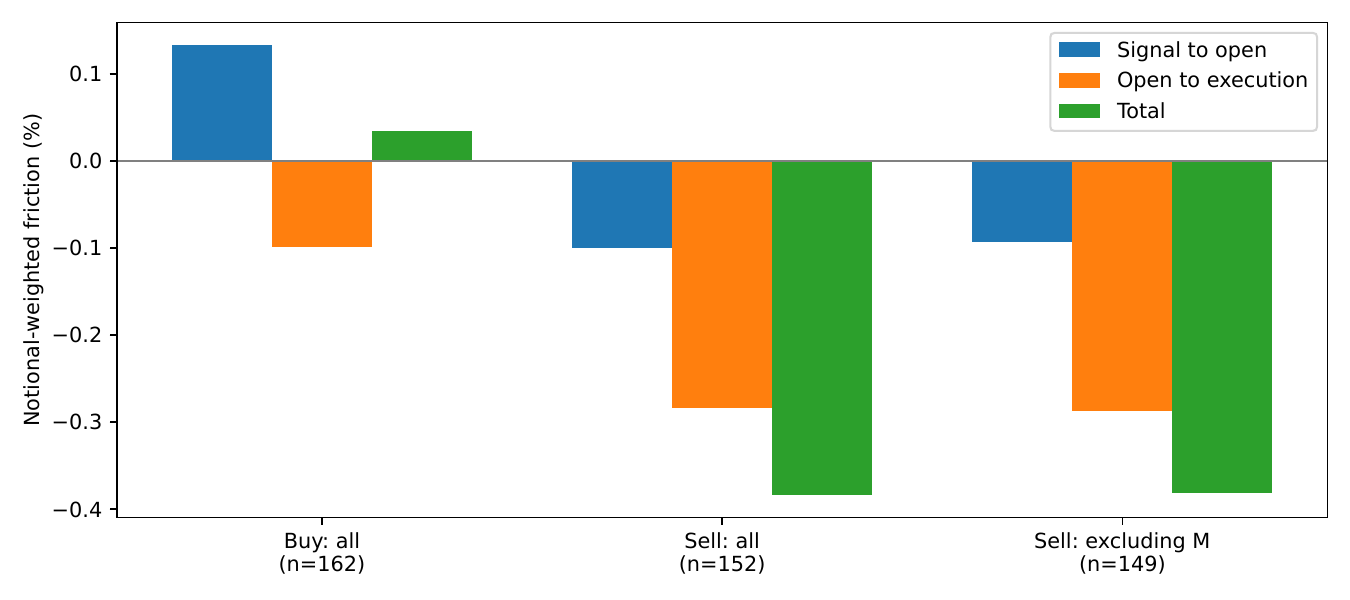}
\caption{Annual transaction-notional-weighted price friction: all 162 Buy fills, all 152 Sell fills, and the 149 Sell fills excluding M. The three M-classified fills are all Sells; there is no separate excluding-M Buy series because it is identical to all Buys. Negative denotes adverse movement; explicit fees are reported separately.}
\label{fig:execution_friction}
\end{figure}

\clearpage
\section{Extended Evidence-to-Claim Matrix}\label{app:evidence-matrix}

Table~\ref{tab:evidence} expands the compact matrix in Table~\ref{tab:claim-main} by retaining the case-specific claims, statuses, and limitations. Its entries describe what the analyzed archive supports, not automatic certification of causal identification.

\begin{longtable}{p{0.29\linewidth}p{0.16\linewidth}p{0.43\linewidth}}
\caption{Extended evidence-to-claim matrix for the analyzed archive.}\label{tab:evidence}\\
\toprule
Claim & Status & Basis / limitation \\
\midrule
\endfirsthead
\toprule
Claim & Status & Basis / limitation \\
\midrule
\endhead
Annual account outcome & Observed & Recorded daily account series, fixed annual period. \\
March loss proves system-specific failure & Not supported & Account outperformed both cited March references despite negative absolute return. \\
April/June T+10 candidate-relative weakness & Supported conditionally & Signal-close anchor, stated candidate scope, approximate gross EW reference. \\
Universal cross-horizon Buy failure & Not supported & Horizon results are mixed across anchors/scopes. \\
Single stock was not causal & Not established & No leave-one-name-out or concentration identification is established. \\
Sell side was irrelevant & Not established & April and June Sell summaries differ materially. \\
Staleness is excluded & Not established & Persistence is recommendation recurrence, not artifact/data age. \\
June is extreme under a strictly matched null & Not supported & The reference is approximate; empirical lower fraction is 5.44\%. \\
June is poor under approximate random ranking & Supported descriptively & 5,000 simulated trials under documented approximations. \\
Drop3 was optimal & Not supported & The replay is non-monotonic; Drop6 exceeds Drop3 on the approximate replay path. \\
May entries caused most actual June loss & Not supported & Account-linked outside group is the largest negative account-linked contribution. \\
One frozen model experienced the full incident & Not established & Factual action bindings $B_a(h)$ are unresolved; config history changed. \\
One recorder bug caused production orders & Not established & Recorded fixes lack action binding and an identified causal comparison. \\
Later strong results prove the same model recovered & Not established & A common frozen runtime identity is not established; candidate evidence is mixed. \\
Complete provenance would prove causality & False as a general rule & Provenance identifies what ran; causal identification needs additional assumptions/evidence. \\
\bottomrule
\end{longtable}

\section{Code and Data Availability}\label{app:availability}

Supplementary Dataset S1 (v1; evidence cutoff 30 September 2026) is available as a separate companion archive on Zenodo at \url{https://doi.org/10.5281/zenodo.23215521}, under the Creative Commons Attribution-ShareAlike 4.0 International license. It contains 241 trading-day returns and 242 closing snapshots including the 30 September 2025 baseline: normalized system, CSI300, and approximate subset-EW NAV/drawdown paths; closing stock counts and set changes; and stable pseudonymous holding memberships. Detailed fields, daily tables, and holding visualizations accompany S1.

S1 summarizes operator-maintained system records, excluding outside-system funds. It supports inspection of account paths and holding-set evolution, not independent brokerage certification or reproduction of every diagnostic table. Stable aliases do not guarantee anonymity. Real security identifiers, capital/position amounts, trade-level records, candidate diagnostics, and private runtime evidence are excluded. Statistical analysis and export scripts remain in the local research archive. The public engine at \href{https://QuantPits.com}{QuantPits.com} is distinct from S1 and is not a complete empirical replication package.


\end{document}